\documentclass{egpubl}
\usepackage{egsr2021}
\usepackage[pagebackref=true,breaklinks=true,colorlinks,bookmarks=false]{hyperref}

\SpecialIssuePaper         

\usepackage[T1]{fontenc}
\usepackage{dfadobe}  
\usepackage{cite}  
\usepackage{times}
\usepackage{graphicx,overpic}
\usepackage{amsmath,amsfonts,amssymb,bm,dsfont,nicefrac}
\usepackage[outline]{contour}
\usepackage{tikz}
\usetikzlibrary{shapes.geometric, arrows}
\usepackage{booktabs}
\usepackage[many]{tcolorbox}
\usepackage{comment}
\usepackage{animate}
\usepackage{enumitem}
\usepackage{egweblnk}

\setitemize{noitemsep,topsep=4pt,parsep=4pt,partopsep=0pt,leftmargin=*}
\setenumerate{noitemsep,topsep=4pt,parsep=4pt,partopsep=0pt,leftmargin=*}

\makeatletter
\renewcommand\paragraph{\@startsection{paragraph}{4}{\z@}%
	{0.6em}%
	{-1em}%
	{\normalfont\normalsize\bfseries}}
\makeatother

\renewcommand{\baselinestretch}{0.99}

\definecolor{fl_color}{rgb}{.0,0.4,0.8}

\definecolor{zd_color}{rgb}{0.8,0.0,0.4}

\definecolor{sz_color}{rgb}{0.0,0.8,0.0}

\newcommand{\revision}[1]{{#1}}

\definecolor{darkgreen}{rgb}{0.05,0.6,0.05}

\newcommand{\D}{\mathrm{d}}
\newcommand{\Real}{\mathbb{R}}

\newcommand{\Le}{I_\mathrm{e}}
\newcommand{\bsdf}{f_\mathrm{r}}
\newcommand{\bp}{\bm{p}}
\newcommand{\bq}{\bm{q}}
\newcommand{\bo}{\bm{o}}
\newcommand{\bx}{\bm{x}}
\newcommand{\by}{\bm{y}}

\newcommand{\mesh}{\mathcal{M}}
\newcommand{\vtx}{\bm{p}}
\newcommand{\uv}{\bm{u}}
\newcommand{\albedoS}{a_\mathrm{s}}
\newcommand{\albedoD}{a_\mathrm{d}}
\newcommand{\roughness}{\alpha}
\newcommand{\norm}{\bm{n}}
\newcommand{\pix}{\mathcal{P}}

\newcommand{\params}{\bm{\xi}}
\newcommand{\Ltot}{\mathcal{L}}
\newcommand{\Lrender}{\mathcal{L}_\mathrm{rend}}
\newcommand{\Lreg}{\mathcal{L}_\mathrm{reg}}
\newcommand{\Lmat}{\mathcal{L}_\mathrm{mat}}

\newcommand{\Lspec}{\mathcal{L}_\mathrm{spec}}
\newcommand{\Lroug}{\mathcal{L}_\mathrm{roug}}
\newcommand{\Lmesh}{\mathcal{L}_\mathrm{mesh}}
\newcommand{\Llap}{\mathcal{L}_\mathrm{lap}}
\newcommand{\Lnorm}{\mathcal{L}_\mathrm{normal}}
\newcommand{\Ledge}{\mathcal{L}_\mathrm{edge}}

\newcommand{\wRender}{\lambda_\mathrm{rend}}

\newcommand{\wSpec}{\lambda_\mathrm{spec}}
\newcommand{\wRoug}{\lambda_\mathrm{roug}}
\newcommand{\wLap}{\lambda_\mathrm{lap}}
\newcommand{\wNorm}{\lambda_\mathrm{normal}}
\newcommand{\wEdge}{\lambda_\mathrm{edge}}

\newcommand{\imgs}{\bm{\mathcal{I}}}
\newcommand{\Irenders}{\imgs}
\newcommand{\Itargets}{\tilde{\imgs}}
\newcommand{\img}{\bm{I}}
\newcommand{\Irender}{\img}
\newcommand{\Itarget}{\tilde{\img}}

\definecolor{blueL}{RGB}{127.5, 177.5, 209.5}
\definecolor{orangeL}{RGB}{251,198,150}
\newcommand{\mymathbox}[3]{%
	\tcboxmath[top=0mm,bottom=0mm,left=0mm,right=0mm,fonttitle=\bfseries\scriptsize\color{gray},colbacktitle=white,enhanced,attach boxed title to top center={yshift=-1mm},boxed title style={top=0mm,bottom=0mm,left=0mm,right=0mm},colframe=#1,colback=white,title=#2]{#3}
}

\newlength{\resLen}

\begin{document}
\BibtexOrBiblatex
\electronicVersion
\title[Unified Shape and SVBRDF Recovery using Differentiable Monte Carlo Rendering]%
{Unified Shape and SVBRDF Recovery\\using Differentiable Monte Carlo Rendering}
%

\author[Luan et al.]
{\parbox{\textwidth}{\centering 
			Fujun Luan$^{1,3}$,
         	Shuang Zhao$^{2}$,
         	Kavita Bala$^{1}$,
         	Zhao Dong$^{3}$
        }
        \\
{\parbox{\textwidth}{\centering $^1$Cornell University\\
         $^2$University of California, Irvine\\
         $^3$Facebook Reality Labs\\
       }
}
}
\newif\ifanimatedteaser
\animatedteaserfalse

\setlength{\resLen}{0.62\columnwidth}
\teaser{
 	\centering
   	\addtolength{\tabcolsep}{-5.6pt}
	\begin{tabular}{cccc}
		\textsc{\scriptsize (a) Capture} &
 		\textsc{\scriptsize (b) Photos} &
 		\textsc{\scriptsize (c) Init. model} &
 		\textsc{\scriptsize (d) Opt. model \hspace{1.5cm} (e) Rendering}
 		\\
 		\includegraphics[height=\resLen]{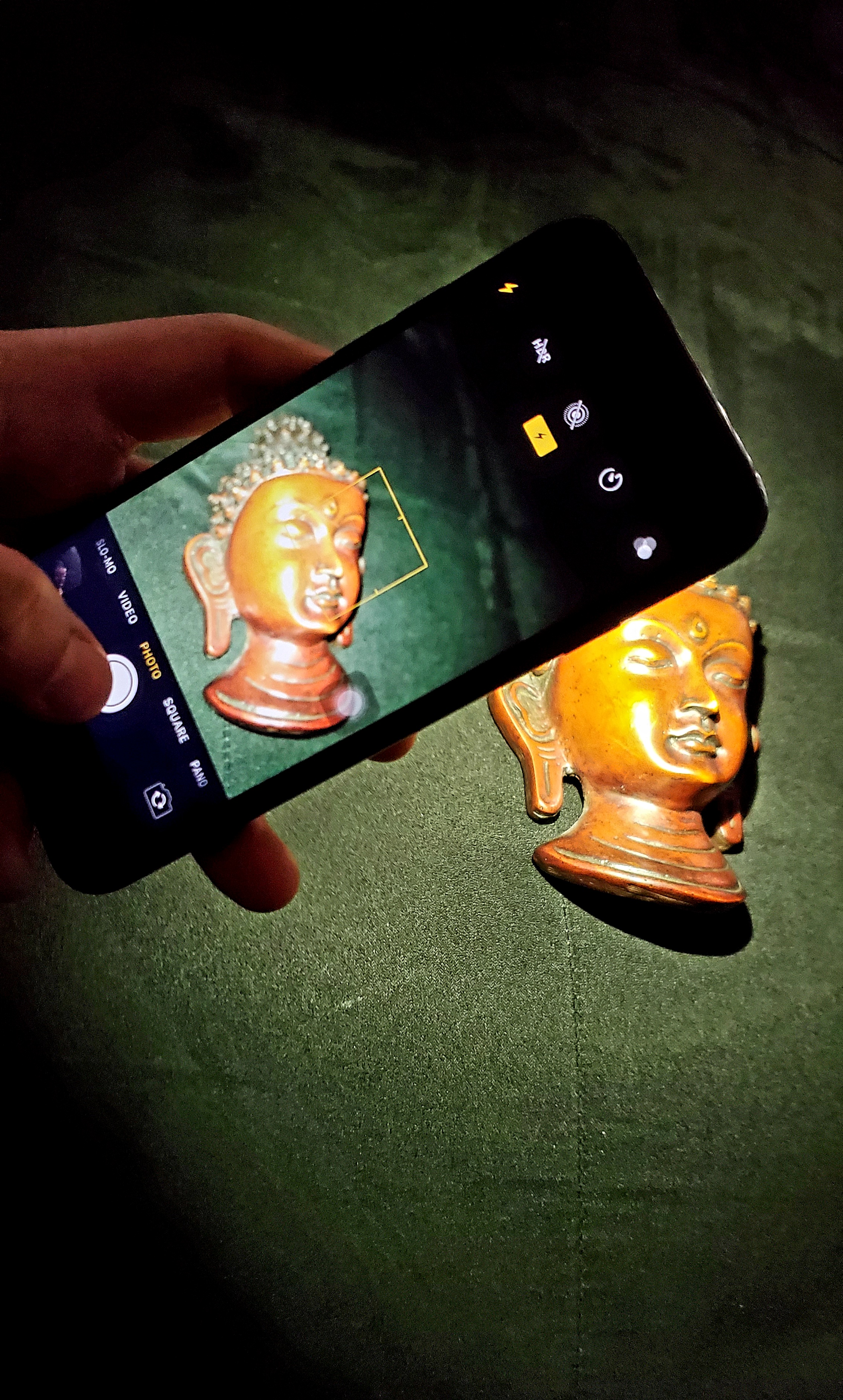} &
 		\includegraphics[height=\resLen]{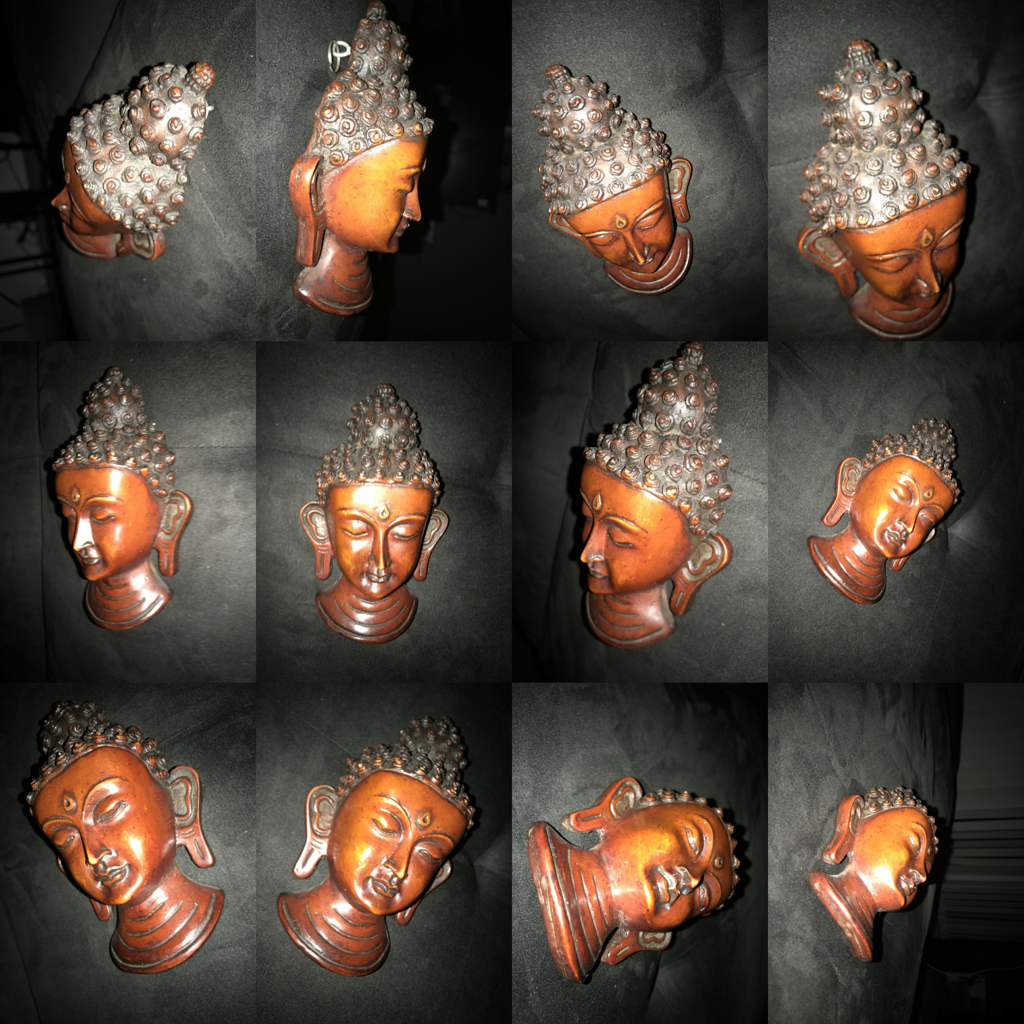} &
 		\includegraphics[height=\resLen]{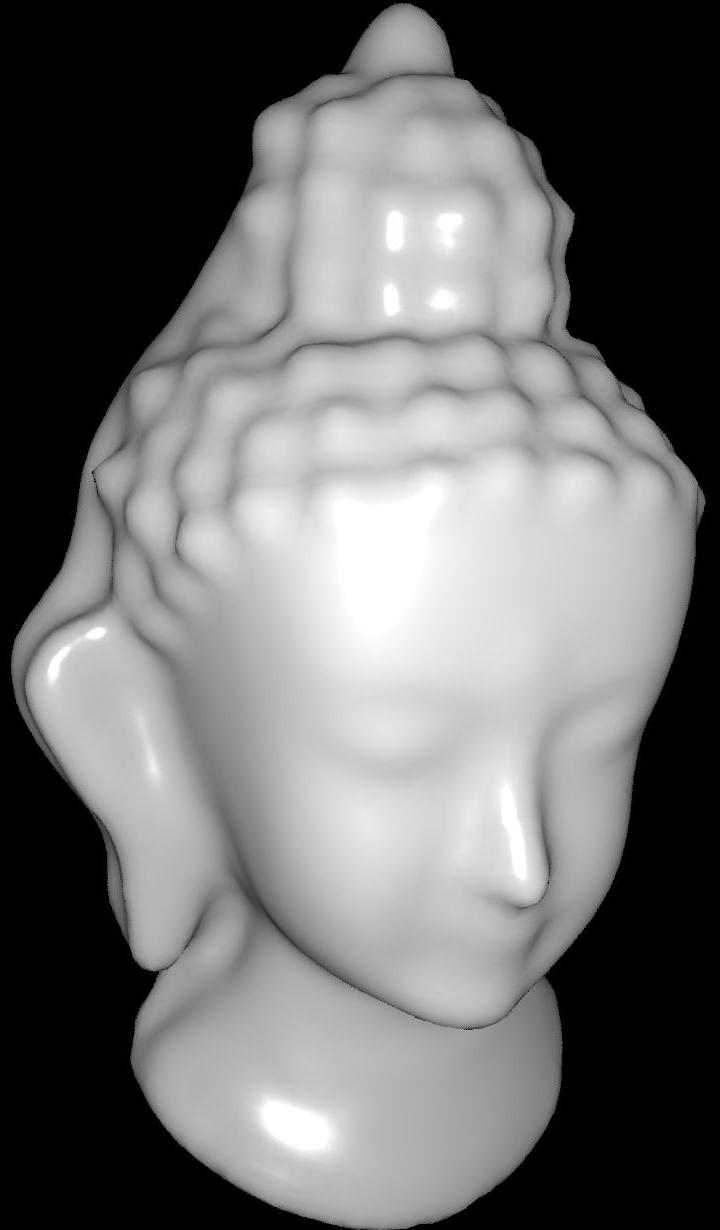} &
 		\ifanimatedteaser
 			\animategraphics[height=\resLen,poster=last,alttext=]{3}{pip/frames/merged_}{0}{19}
 		\else
 			\includegraphics[height=\resLen]{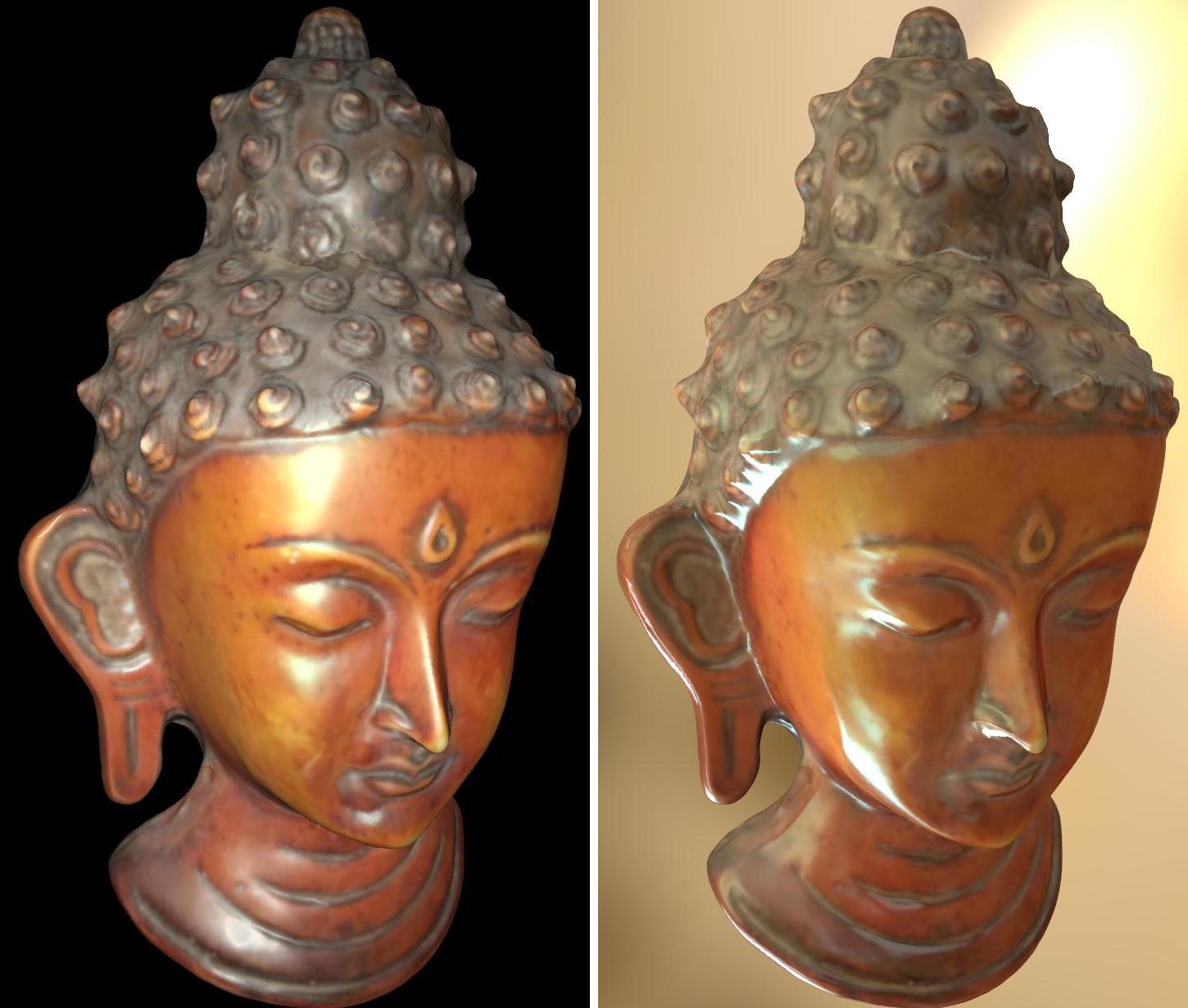}
 		\fi
	\end{tabular}
 	\caption{\label{fig:pip}
		(a) Capture setup. Our method takes as input: multi-view photos (100 captured and 12 shown for this example) of an object~(b) and a rough initial model with its geometry obtained using standard methods~(c). Then, a novel analysis-by-synthesis optimization is performed to refine the model's shape and reflectance in a unified fashion, yielding a high-quality 3D model~(d). We show in (e) a re-rendering of the result under environmental lighting. 
		\ifanimatedteaser
			\emph{(Please use Adobe Acrobat and click on (d) or (e) to see an animation of the optimization process.)}
		\fi
	}
}

\maketitle
\begin{abstract}
	Reconstructing the shape and appearance of real-world objects using measured 2D images has been a long-standing inverse rendering problem.
	In this paper, we introduce a new analysis-by-synthesis technique capable of producing high-quality reconstructions through robust coarse-to-fine optimization and physics-based differentiable rendering.

	Unlike most previous methods that handle geometry and reflectance largely separately, our method unifies the optimization of both by leveraging image gradients with respect to both object reflectance {\em and} geometry.	
	To obtain physically accurate gradient estimates, we develop a new GPU-based Monte Carlo differentiable renderer leveraging recent advances in differentiable rendering theory to offer unbiased gradients while enjoying better performance than existing tools like PyTorch3D~\cite{ravi2020accelerating} and redner~\cite{li2018differentiable}.
	To further improve robustness, we utilize several shape and material priors as well as a coarse-to-fine optimization strategy to reconstruct geometry.
	Using both synthetic and real input images, we demonstrate that our technique can produce reconstructions with higher quality than previous methods. 
\end{abstract}

\section{Introduction}
\label{sec:intro}
Reconstructing the shape and appearance of real-world objects from 2D images has been a long-standing problem in computer vision and graphics.
Previously, the acquisition of object geometry and (spatially varying) reflectance has been studied largely independently.
For instance, many techniques based on multiview-stereo (MVS)~\cite{ghosh2008practical,schwartz2013dome,tunwattanapong2013acquiring, nam2016simultaneous,aittala2015two, hui2017reflectance, riviere2016mobile, riviere2017polarization} and time-of-flight imaging~\cite{newcombe2011kinectfusion, Izadi2011KinectFusion} have been introduced for the reconstruction of 3D shapes.
Although these methods can also provide rough estimations of surface reflectance, they usually rely on the assumption of simple (e.g., diffuse-dominated) reflectance and can produce unsatisfactory results for glossy objects.
On the other hand, previous approaches that specialized at recovering an object's spatially varying reflectance~\cite{zhou2016sparse,gao2019deep,Guo:2020:MaterialGAN} typically require object geometries to be predetermined, limiting their practical usage for many applications where such information is unavailable.

\revision{
Recently, great progress has been made in the area of Monte Carlo differentiable rendering. On the other hand, how this powerful tool can be applied to solve practical 3D reconstruction problems---a main application area of differentiable rendering---has remained largely overlooked. Prior works (e.g.,~\cite{nam2018practical}) have mostly relied on alternative Poisson reconstruction steps during the optimization, leading to suboptimal geometry quality. Instead, by leveraging edge sampling that provides unbiased gradients of mesh vertex positions, we optimize object shape and SVBRDF in a unified fashion, achieving state-of-the-art reconstruction quality. 
}

In this paper, we demonstrate that detailed geometry and spatially varying reflectance of a real-world object can be recovered using a unified \emph{analysis-by-synthesis} framework.
To this end, we apply gradient-based optimization of the rendering loss (i.e., the difference between rendered and target images) that are affected by both object geometry and reflectance.
Although such gradients with respect to appearance are relatively easy to compute, the geometric gradients are known to be much more challenging to compute and, therefore, have been mostly approximated in the past using techniques like soft rasterization~\cite{liu2019soft} in computer vision.
We, on the other hand, leverage recent advances in physics-based differentiable rendering to obtain \emph{unbiased} and \emph{consistent} geometric gradients that are crucial for obtaining high-quality reconstructions.

Concretely, our contributions include:
\begin{itemize}
	\item A Monte Carlo differentiable renderer specialized for collocated configurations. Utilizing edge sampling~\cite{li2018differentiable}, our renderer produces unbiased and consistent gradient estimates.
    \item A new analysis-by-synthesis pipeline that enables high-quality reconstruction of spatially varying reflectance and, more importantly, mesh-based object geometry.
    \item A coarse-to-fine scheme as well as geometry and reflectance priors for ensuring robust reconstructions.
    \item \revision{Thorough validations and evaluations of individual steps that come together allowing practical and high-quality 3D reconstruction using inexpensive handheld acquisition setups, which benefits applications in many areas like graphics and AR/VR.}
\end{itemize}
%
We demonstrate the effectiveness of our technique via several synthetic and real examples.


\section{Related work}
\label{sec:related}
\paragraph*{Shape reconstruction.}
Reconstructing object geometry has been a long-standing problem in computer vision.

\emph{Multi-view Stereo~(MVS)} recovers the 3D geometry of sufficiently textured objects using multiple images of an object by matching feature correspondences across views and optimizing photo-consistency (e.g., \cite{seitz1999photorealistic, vogiatzis2005multi, furukawa2009accurate,schoenberger2016mvs}).

\emph{Shape from Shading~(SfS)} relates surface normals to image intensities~\cite{horn1970shape,ikeuchi1981numerical,queau2017variational,queau2017dense,maier2017intrinsic3d,haefner2018fight}.
Unfortunately, these methods have difficulties handling illumination changes, non-diffuse reflectance, and textureless surfaces.

\emph{Photometric Stereo~(PS)} takes three or more images captured with a static camera and varying illumination or object pose, and directly estimate surface normals from measurements~\cite{woodham1980photometric,holroyd2008photometric,zhou2010ring,tunwattanapong2013acquiring, papadhimitri2014uncalibrated,queau2015solving,queau2016unbiased}.
These methods typically do not recover reflectance properties beyond diffuse albedo.

\paragraph*{Reflectance reconstruction.}
Real-world objects exhibit richly diverse reflectance that can be described with spatially-varying bidirectional reflectance distribution functions (SVBRDFs).

Traditional SVBRDF acquisition techniques rely on dense input images measured using light stages or gantry (e.g, \cite{matusik2003data,lensch2003image,holroyd2010coaxial,dong2010manifold,chen2014reflectance,dong2015predicting,kang2018efficient}).
To democratize the acquisition, some recent works exploit the structure (e.g., sparsity) of SVBRDF parameter spaces to allow reconstructions using fewer input images (e.g., \cite{yu1999inverse,dong2014appearance,wu2015simultaneous,zhou2016sparse,kim2017lightweight,park2018surface}). Additionally, a few recent works have been introduced to produce plausible SVBRDF estimations for flat objects using a small number of input images (e.g., \cite{aittala2015two,aittala2016reflectance,hui2017reflectance,gao2019deep,deschaintre2019flexible,Guo:2020:MaterialGAN}).
Despite their ease of use, these techniques cannot be easily generalized to handle more complex shapes.


\paragraph*{Joint estimation of shape and reflectance.}
%
Several prior works jointly estimate object shape and reflectance.
Higo et al.~\cite{higo2009hand} presented a plane-sweeping method for albedo, normal and depth estimation.
Xia et al.~\cite{xia2016recovering} optimized an apparent normal field with corresponding reflectance.
Nam et al.~\cite{nam2018practical} proposed a technique that alternates between material-, normal-, and geometry-optimization stages.
Schmitt et al.~\cite{schmitt2020joint} 
perform joint estimation using a hand-held sensor rig with 
and 12 point light sources.
Bi et al.~\cite{bi2020deep} use six images 
and optimize object geometry and reflectance in two separate stages.

All these methods either rely on MVS for geometry reconstruction or perform alternative optimization of shape and reflectance, offering little to no guarantee on qualities of the reconstruction results.
We, in contrast, formulate the problem as a unified analysis-by-synthesis optimization, ensuring locally optimal results.

\paragraph*{Differentiable rendering of meshes.}
We now briefly review differentiable rendering techniques closely related to our work.
For a more comprehensive summary, please see the survey by Kato~et~al.~\cite{kato2020differentiable}.

Specialized differentiable renderers have long existed in computer graphics and vision~\cite{gkioulekas2013inverse,gkioulekas2016evaluation,tsai2019beyond,azinovic2019inverse,che2020towards}.
Recently, several general-purpose ones~\cite{li2018differentiable,nimier2019mitsuba} have been developed.

A key technical challenge in differentiable rendering is to estimate gradients with respect to object geometry (e.g., positions of mesh vertices).
To this end, several approximated methods (e.g., \cite{loubet2019reparameterizing,liu2019soft,ravi2020accelerating}) have been proposed.
Unfortunately, inaccuracies introduced by these techniques can lead to degraded result quality.
On the contrary, Monte Carlo edge sampling~\cite{li2018differentiable,Zhang:2019:DTRT}, which we use for our differentiable renderer, provides unbiased gradient estimates capable of producing higher-quality reconstructions.

\section{Our method}
\label{sec:our_method}
We formulate the problem of joint estimation of object geometry and reflectance as an \emph{analysis-by-synthesis} (aka. inverse-rendering) optimization.
Let $\params$ be some vector that depicts both the geometry and the reflectance of a real-world object.
Taking as input a set of images $\Itargets$ of this object, we estimate $\params$ by minimizing a predefined loss $\Ltot$:
\begin{equation}
	\label{eq:inv_render}
	\params^* = 
	\textstyle\arg\min_{\params} \Ltot (\Irenders(\params), \params;\;\Itargets),
\end{equation}
where $\Irenders(\params)$ are a set of renderings of the object generated using the geometry and reflectance provided by $\params$.
We further allow the loss $\Ltot$ to directly depend on the parameters $\params$ for regularization.

\paragraph*{Acquisition setup.}
Similar to recent works on reflectance capture~\cite{aittala2015two, albert2018approximate, hui2017reflectance, riviere2016mobile, aittala2016reflectance, gao2019deep, deschaintre2019flexible, ren2011pocket}, we utilize an acquisition setup where the object is illuminated with a point light collocated with the camera. 
\revision{
	This collocated configuration significantly simplifies both forward and differentiable rendering processes, allowing the analysis-by-synthesis problem to be solved efficiently.
}
Common \revision{collocated configurations} include a smartphone's flash and camera as well as a consumer-grade RGBD sensor mounted with an LED light.

\paragraph*{Overview of our method.}
Efficiently solving the optimization of Eq.~\eqref{eq:inv_render} requires computing gradient $\nicefrac{\D\Ltot}{\D\params}$ of the loss $\Ltot$ with respect to the geometry and reflectance parameters $\params$.
According to the chain rule, we know that
\begin{equation}
	\label{eq:inv_render_grad}
	\frac{\D \Ltot}{\D \params} = \frac{\partial\Ltot}{\partial\Irenders} \frac{\D\Irenders}{\D\params} + \frac{\partial\Ltot}{\partial\params},
\end{equation}
where $\nicefrac{\partial\Ltot}{\partial\Irenders}$ and $\nicefrac{\partial\Ltot}{\partial\params}$ can be computed using automatic differentiation~\cite{paszke2017automatic}.
Further, estimating gradients~$\nicefrac{\D\Irenders}{\D\params}$ of rendered images requires performing differentiable rendering.
Despite being relatively easy when the parameters $\params$ only capture reflectance, differentiating the rendering function $\Irenders$ becomes much more challenging when $\params$ also controls object geometry~\cite{li2018differentiable}.
To this end, we develop a new differentiable renderer that is specific to our acquisition setup and provides unbiased gradient estimates.

In the rest of this section, we provide a detailed description of our technique that solves the analysis-by-synthesis optimization~\eqref{eq:inv_render} in an efficient and robust fashion.
In \S\ref{ssec:pbdr}, we detail our forward-rendering model and explain how it can be differentiated.
In \S\ref{ssec:opt}, we discuss our choice of the loss $\Ltot$ and optimization strategy.

\begin{figure}[t]
	\centering
	\includegraphics[width=.995\columnwidth]{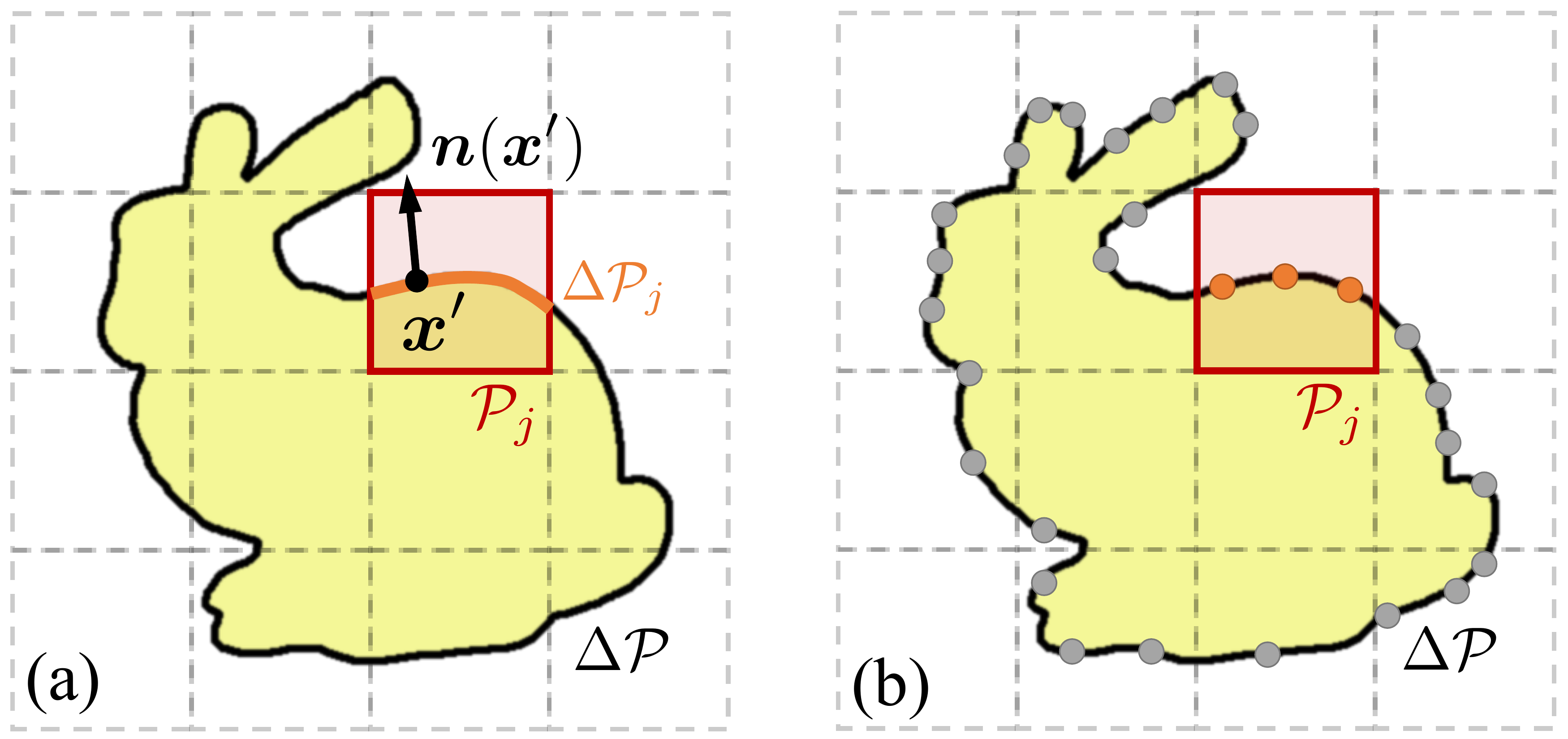}
	\caption{\label{fig:diff_render}
		\textbf{Differentiable rendering:}
		(a) To properly differentiate the intensity $I_j$ of a pixel $\pix_j$ (illustrated as red squares) with respect to object geometry, a boundary integral over $\Delta\pix_j$ (illustrated as the orange curve) needs to be calculated.
		(b) We perform Monte Carlo edge sampling~\cite{li2018differentiable,Zhang:2019:DTRT} by (i)~sampling points (illustrated as small discs) from pre-computed discontinuity curves~$\Delta\pix$, and (ii)~accumulating their contributions in the corresponding pixels (e.g., the orange samples contribute to the pixel $\pix_j$).
	}
\end{figure}

\begin{figure*}[t]
	\centering
    \setlength{\resLen}{0.94in}
	\newcommand{\ablationOne}[1]{%
		\includegraphics[width=\resLen]{results/ablation1/#1/init3.jpg} &
		\includegraphics[width=\resLen]{results/ablation1/#1/softras3.jpg} &
		\includegraphics[width=\resLen]{results/ablation1/#1/pytorch3d3.jpg} &
		\includegraphics[width=\resLen]{results/ablation1/#1/mitsuba3.jpg} &
		\includegraphics[width=\resLen]{results/ablation1/#1/nvdiffrast3.jpg} &
		\includegraphics[width=\resLen]{results/ablation1/#1/ours3.jpg} &
		\includegraphics[width=\resLen]{results/ablation1/#1/gt2.jpg}%
	}
	\addtolength{\tabcolsep}{-4pt}
	\small
   	\begin{tabular}{ccccccc}
		\textsc{\scriptsize Init. mesh} &
		\textsc{\scriptsize SoftRas} &
		\textsc{\scriptsize Pytorch3D} &
		\textsc{\scriptsize Mitsuba 2} &
		\textsc{\scriptsize Nvdiffrast} &
		\textsc{\scriptsize Ours} &
		\textsc{\scriptsize Ground truth}\\
		\ablationOne{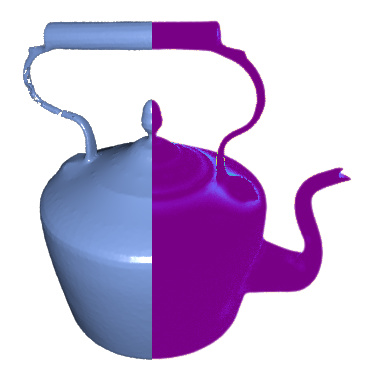}\\[-4pt]
		0.0411 & 0.0051 & 0.0072 & 0.0071 & 0.0013 &\textbf{0.0004} & \textit{kettle}\\
		\ablationOne{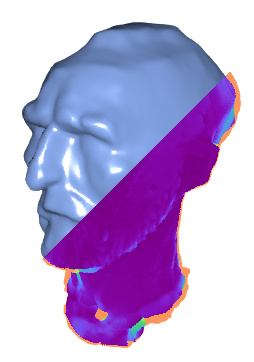}\\[-4pt]
		0.0115 & 0.0039 & 0.0091 & 0.0065 & 0.0022 &\textbf{0.0016} & \textit{head}\\
		\ablationOne{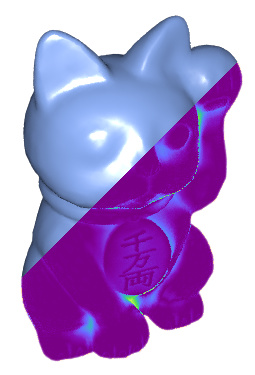}\\[-4pt]
		0.0878 & 0.0053 & 0.0066 & 0.0071 & 0.0023 &\textbf{0.0010} & \textit{maneki}\\[-2pt]
	\end{tabular}
	\includegraphics[width=0.475\textwidth]{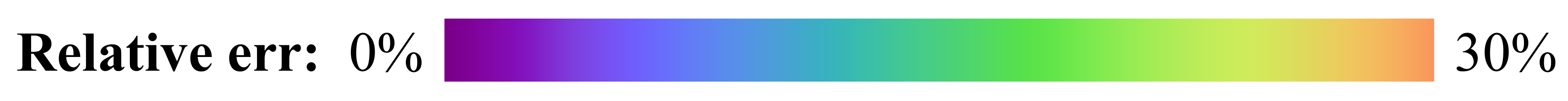}
    \caption{\label{fig:ablation1}
    	\textbf{Comparison} with SoftRas~\cite{liu2019soft}, PyTorch3D~\cite{ravi2020accelerating}, Mitsuba 2~\cite{nimier2019mitsuba} and Nvdiffrast~\cite{Laine2020diffrast}.
    	We render all reconstructed geometries using Phong shading and visualize depth errors (wrt. the ground-truth geometry).
    	Initialized with the same mesh (shown in the left column), optimizations using gradients obtained with SoftRas and PyTorch3D tend to converge to low-quality results due to gradient inaccuracies caused by soft rasterization.
    	Mitsuba~2, a ray-tracing-based system, also produces visible artifacts due to biased gradients resulting from an approximated reparameterization~\cite{loubet2019reparameterizing}. Nvdiffrast is using multisample analytic antialiasing method to provide reliable visibility gradients, which yields better optimization result overall. 
    	When using gradients generated with our differentiable renderer, optimizations under identical configurations produce results closely resembling the targets.
    	The number below each result indicates the average point-to-mesh distance capturing the Euclidean accuracy~\cite{jensen2014large} of the reconstructed geometry (normalized to have a unit bounding box).
    }
    
\end{figure*}

\begin{figure*}[t]
 	\centering
 	\small
  	\setlength{\resLen}{1.1in}
	\addtolength{\tabcolsep}{-4pt}
	\begin{tabular}{ccccccc}
		\textsc{\scriptsize Init. mesh} &
		\textsc{\scriptsize No Laplacian} &
		\textsc{\scriptsize No normal/edge loss } &
		\textsc{\scriptsize No coarse-to-fine} &
		\textsc{\scriptsize Ours} &
		\textsc{\scriptsize Ground truth}\\
		\includegraphics[width=\resLen]{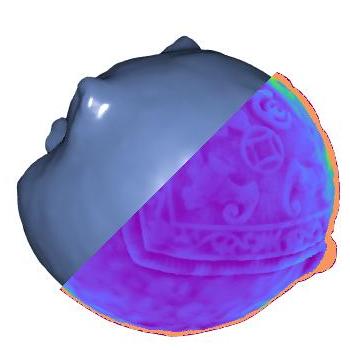} &
		\includegraphics[width=\resLen]{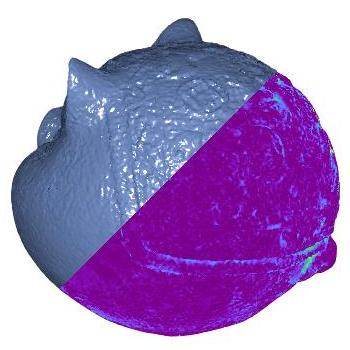} &
		\includegraphics[width=\resLen]{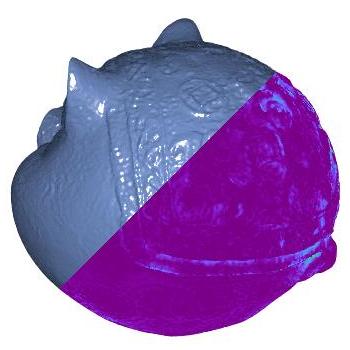} &
		\includegraphics[width=\resLen]{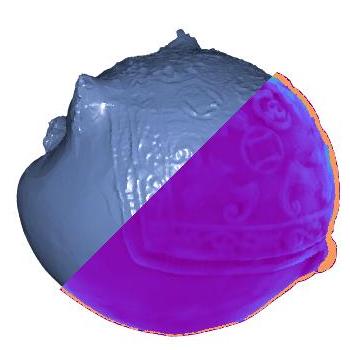} &
		\includegraphics[width=\resLen]{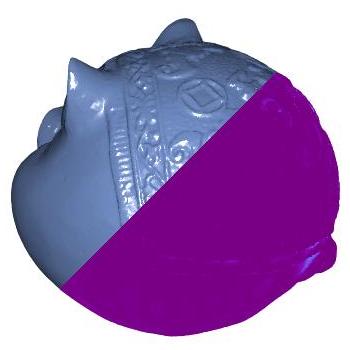} &
		\includegraphics[width=\resLen]{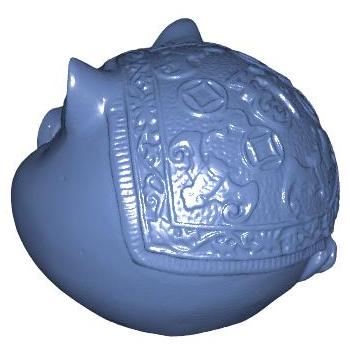}\\[-3pt]
		 0.0318 & 0.0014 & 0.0013 & 0.0137 & \textbf{0.0006} & \textit{pig}\\[2pt]
		\includegraphics[width=\resLen]{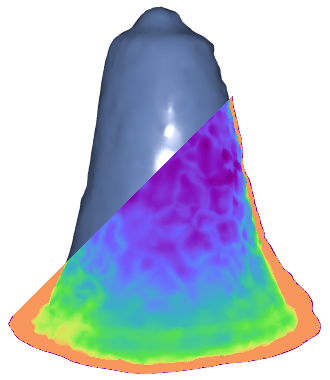} &
		\includegraphics[width=\resLen]{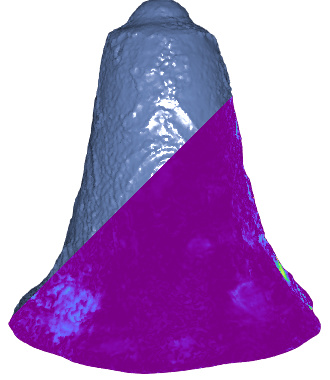} &
		\includegraphics[width=\resLen]{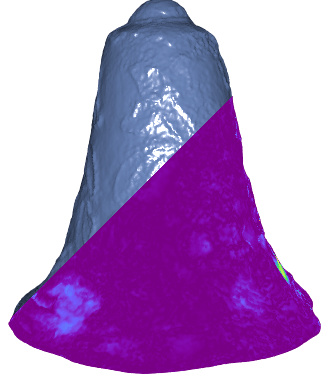} &
		\includegraphics[width=\resLen]{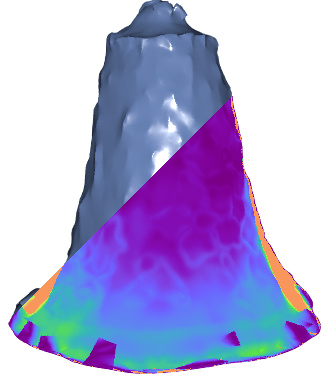} &
		\includegraphics[width=\resLen]{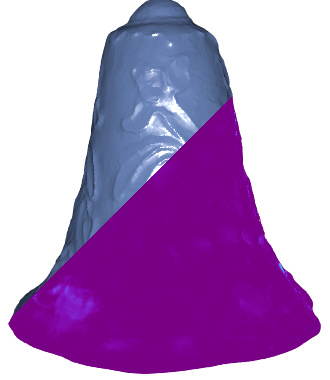} &
		\includegraphics[width=\resLen]{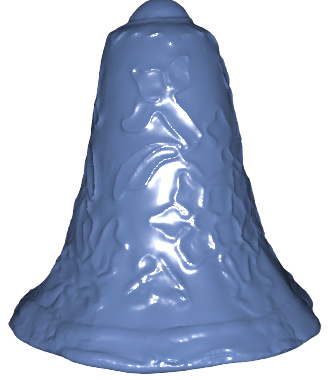}\\[-3pt]
		 0.0197 & 0.0011 & 0.0010 & 0.0132 & \textbf{0.0005} & \textit{bell}\\[2pt]
   	\end{tabular}
    \caption{\label{fig:ablation3}
    	\textbf{Ablation study} on our mesh loss of Eq.~\protect\eqref{eq:los_reg_mesh} and coarse-to-fine framework.
    	Using the identical initializations and optimization settings, we show geometries (rendered under a novel view) optimized with (i)~various components of the mesh loss; and (ii)~the coarse-to-fine process disabled.
    	Similar to Figure~\protect\ref{fig:ablation1}, the number below each result indicates the average point-to-mesh distance.
   }
\end{figure*}

\subsection{Forward and differentiable rendering}
\label{ssec:pbdr}
In what follows, we describe (i)~our representation of object geometry and reflectance; and (ii)~how we render these representations in a differentiable fashion.

\paragraph*{Object geometry and reflectance.}
We express object geometries using standard triangle meshes.
Compared to other representations that are popular in 3D reconstruction, such as SDF volumes~\cite{park2019deepsdf,jiang2020sdfdiff,physg2020}, occupancy networks~\cite{mescheder2019occupancy} or sphere-based clouds~\cite{lassner2020fast}, triangle meshes can be efficiently rendered and edited with many 3D digital content creation tools.
Further, as a widely adopted format, triangle meshes can be easily imported into numerous applications in computer graphics, vision, and augmented/virtual reality (AR/VR).

One of the biggest challenges when using meshes for 3D reconstruction is that topological changes are difficult. We show in the following sections that this can be addressed by using reasonable initial geometries and a coarse-to-fine optimization process.

To depict an object's spatially varying reflectance, we use the Disney BRDF~\cite{karis2013real}, a parametric model offering a good balance between simplicity and flexibility.
This model has also been used by many prior works (e.g., \cite{li2018materials,li2020inverse,bi2020deep}).
Using this BRDF model, the spatially varying reflectance of an object is described using \revision{three} 2D texture maps specifying, respectively, diffuse albedo~$\albedoD$, specular albedo~$\albedoS$, surface roughness~$\roughness$. \revision{And surface normals~$\norm$ are computed from updated mesh vertex positions at every step}. 
Thanks to the efficiency of our system (which we will present in the following), we directly use fine meshes to express detailed geometries and do not rely on approximations like bump/normal mapping.

\paragraph*{Forward rendering.}
Given a virtual object depicted using parameters $\params$, we render one-bounce reflection (aka. direct illumination) of the object.
Specifically, assume the point light and the camera are collocated at some $\bo \in \Real^3$.
Then, the intensity $I_j$ of the $j$-pixel is given by an area integral over the pixel's footprint $\pix_j$, which is typically a square on the image plane:
\begin{equation}
	\label{eq:render}
	I_j = \frac{1}{| \pix_j |} \int_{\pix_j}  \underbrace{\Le\,\frac{\bsdf(\bo \to \by \to \bo)}{\| \by - \bo \|^2}}_{=:\,I(\bx)} \,\D A(\bx),
\end{equation}
where $\by$ is the intersection between the object geometry $\mesh$ and a ray that originates at $\bo$ and passes through $\bx$ on the image plane.
Further, $\Le$ denotes the intensity of the point light; $\bsdf(\bo \to \by \to \bo)$ indicates the cosine-weighted BRDF at $\by$ (evaluated with both the incident and the outgoing directions pointing toward $\bo$); and $A$ is the surface-area measure.
%
We note that no visibility check is needed in Eq.~\eqref{eq:render} since, under the collocated configuration, any point~$\by \in \mesh$ visible to the camera must be also visible to the light source.

We estimate Eq.~\eqref{eq:render} using Monte Carlo integration by uniformly sampling $N$ locations $\bx_1, \bx_2, \ldots, \bx_N \in \pix_j$ and computing $I_j \approx \frac{1}{N} \sum_{i = 1}^N I(\bx_i)$ where $I$ is the integrand defined in Eq.~\eqref{eq:render}.

\paragraph*{Differentiable rendering.}
%
Computing image gradients $\nicefrac{\D\Irenders}{\D\params}$ in Eq.~\eqref{eq:inv_render_grad} largely boils down to differentiating pixel intensities Eq.~\eqref{eq:render} with respect to $\params$.
Although this can sometimes be done by differentiating the integrand $I$---that is, by estimating $\int_{\pix_j} (\nicefrac{\D I}{\D\params}) \,\D A$---doing so is insufficient when computing gradients with respect to object geometry (e.g., vertex positions). 
Consequently, the gradient~$\nicefrac{\D\Irenders}{\D\params}$ has usually been approximated using soft rasterization~\cite{liu2019soft,ravi2020accelerating} or reparameterized integrals~\cite{loubet2019reparameterizing}.
Biased gradient estimates, unfortunately, can reduce the quality of optimization results, which we will demonstrate in \S\ref{sec:results}.

On the other hand, a few general-purpose unbiased techniques~\cite{li2018differentiable,Zhang:2019:DTRT} have been introduced recently.
Unfortunately, these methods focus on configurations without point light sources---which is not the case under our collocated configuration.
We, therefore, derive the gradient $\nicefrac{\D I_j}{\D\params}$ utilizing mathematical tools used by these works.
Specifically, according to Reynolds transport theorem~\cite{reynolds1903papers}, the gradient involves an \emph{interior} and a \emph{boundary} integrals: 
\begin{align}
	\label{eq:drender} 
	\frac{\D I_j}{\D\params} = \tfrac{1}{| \pix_j |} \bigg[ & \; \mymathbox{blueL}{interior}{\textstyle\int_{\pix_j}  \frac{\D I}{\D\params}(\bx) \,\D A(\bx)} \;+\\
	\nonumber
	& \; \mymathbox{orangeL}{boundary}{\textstyle\int_{\Delta\pix_j} \left(\norm(\bx') \cdot \frac{\D\bx'}{\D\params}\right) \Delta I(\bx') \,\D\ell(\bx')} \;\bigg],
\end{align}
where the interior term is simply Eq.~\eqref{eq:render} with its integrand~$I$ differentiated.
The boundary one, on the contrary, is over curves $\Delta\pix_j := \Delta\pix \cap \pix_j$ with $\Delta\pix$ comprised of jump discontinuity points of $I$.
In practice, $\Delta\pix$ consists of image-plane projections of the object's silhouettes.
Further, $\norm(\bx)$ is the curve normal within the image plane, $\Delta I(\bx)$ denotes the difference in $I$ across discontinuity boundaries, and $\ell$ is the curve-length measure (see Figure~\ref{fig:diff_render}-a).

Similar to the Monte Carlo estimation of Eq.~\eqref{eq:render}, we estimate the interior integral in Eq.~\eqref{eq:drender} by uniformly sampling $\bx_1, \ldots, \bx_N \in \pix_j$.
To handle the boundary integral, we precompute the discontinuity curves $\Delta\pix$ (as polylines) by projecting the object's silhouette onto the image plane \revision{at each iteration}.
At runtime, we draw $\bx'_1, \ldots, \bx'_M \in \Delta\pix$ uniformly.
Then,
\begin{align}
	\label{eq:drender1}
	\frac{\D I_j}{\D\params} \approx\; & \mymathbox{blueL}{interior}{\textstyle\frac{1}{N} \sum_{i = 1}^{N} \frac{\D I}{\D\params}(\bx_i)} \;+\\
	\nonumber
	& \mymathbox{orangeL}{boundary}{\textstyle\frac{1}{M} \frac{|\Delta\pix|}{|\pix_j|} \sum_{i = 1}^{M} \mathds{1}[\bx'_i \in \pix_j] \left(\norm(\bx'_i) \cdot \frac{\D\bx'_i}{\D\params}\right) \Delta I(\bx'_i)},
\end{align}
where $|\Delta\pix|$ denotes the total length of the discontinuity curves~$\Delta\pix$, and $\mathds{1}[\cdot]$ is the indicator function.

In practice, we estimate gradients of pixel intensities via Eq.~\eqref{eq:drender1} in two rendering passes.
In the first pass, we evaluate the interior component independently for each pixel.
In the second pass, we evaluate the boundary component for each $\bx'_i$ in parallel and accumulate the results in the corresponding pixel (see Figure~\ref{fig:diff_render}-b).

%
\begin{figure*}[t]
 	\centering
 	\small
  	\setlength{\resLen}{1.05in}
	\addtolength{\tabcolsep}{-4.5pt}
	\begin{tabular}{cccc}
		\textsc{\scriptsize No spec. correlation} &
		\textsc{\scriptsize No roug. smoothness } &
		\textsc{\scriptsize Ours} &
		\textsc{\scriptsize Ground truth}\\
		\includegraphics[height=\resLen]{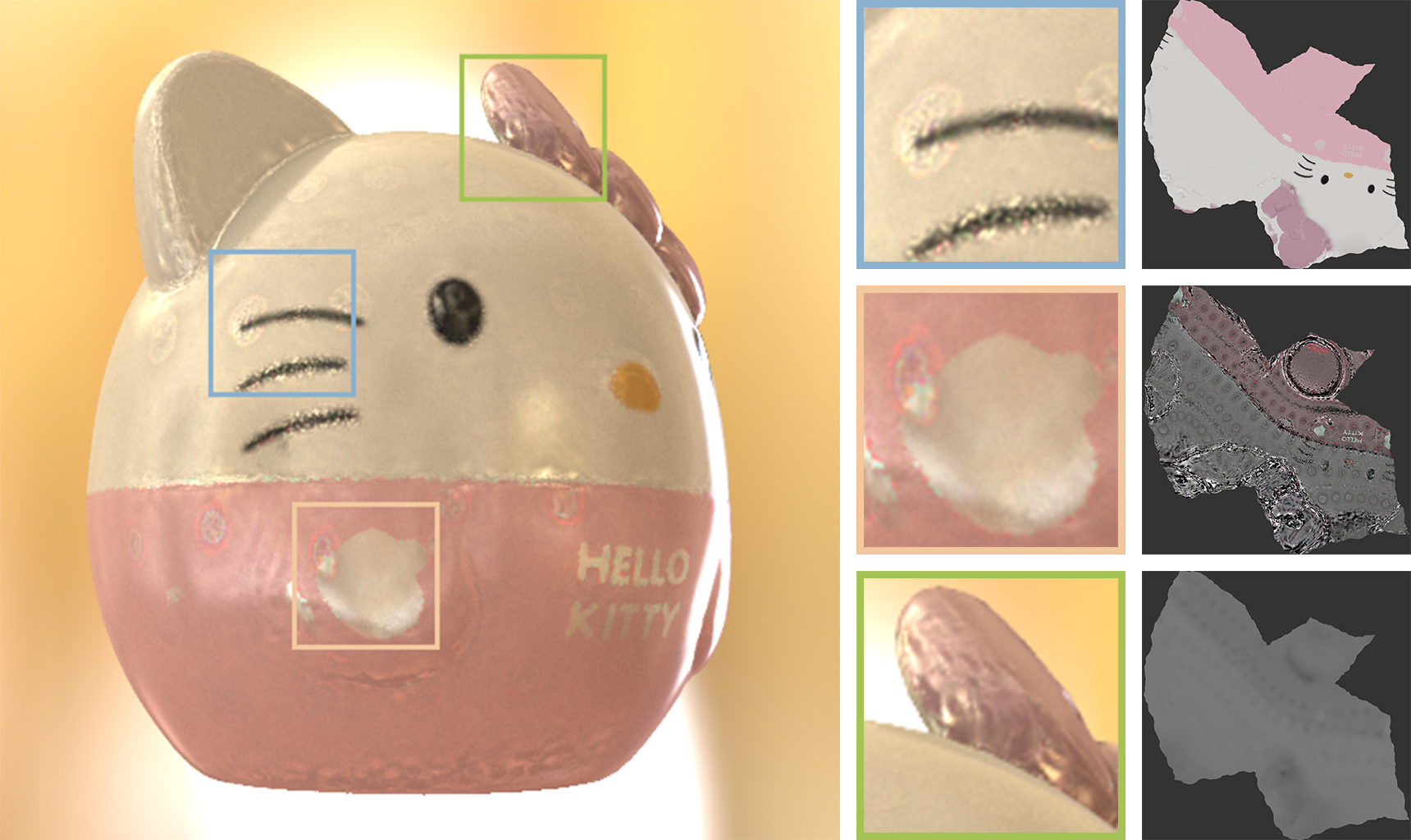} &
		\includegraphics[height=\resLen]{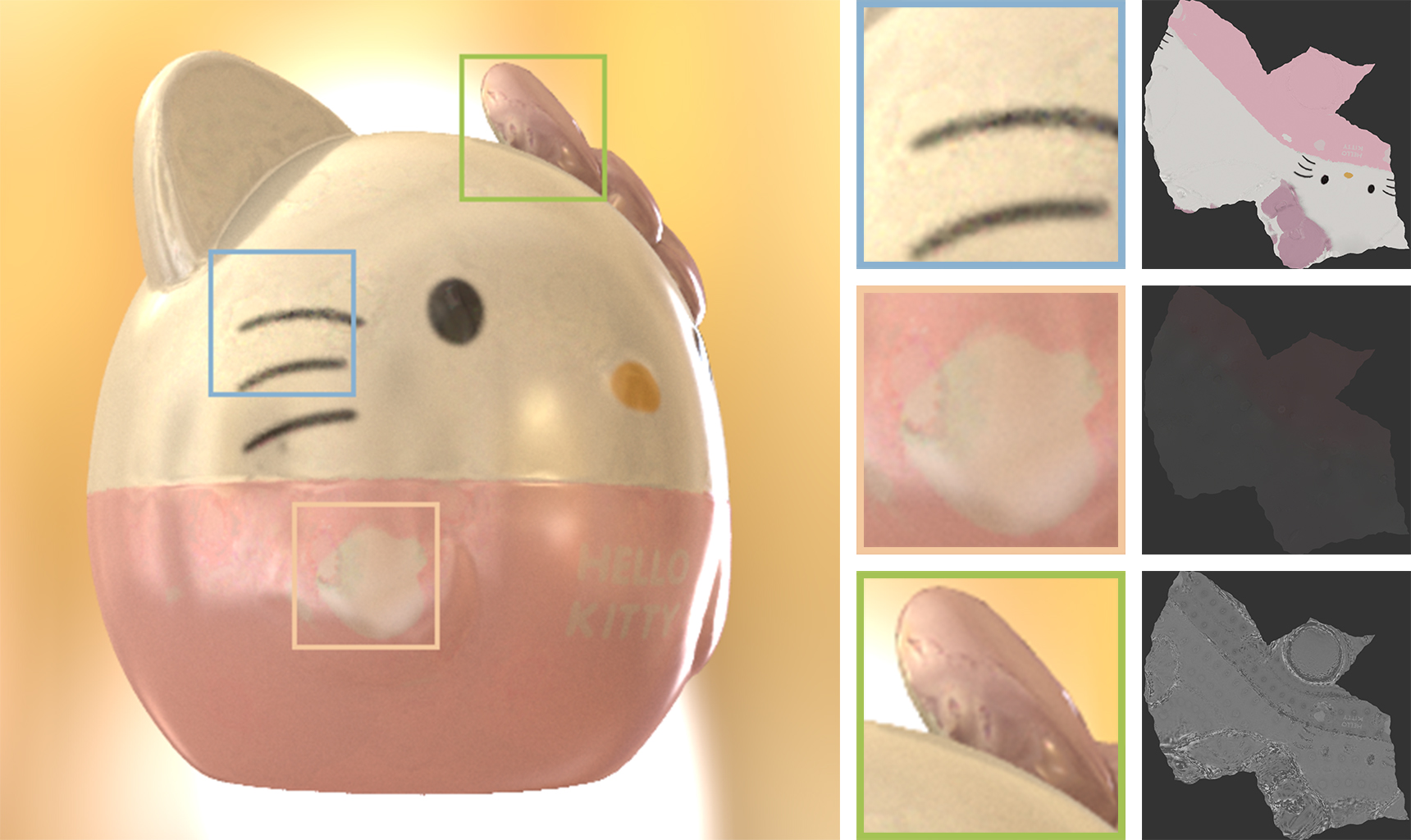} &
		\includegraphics[height=\resLen]{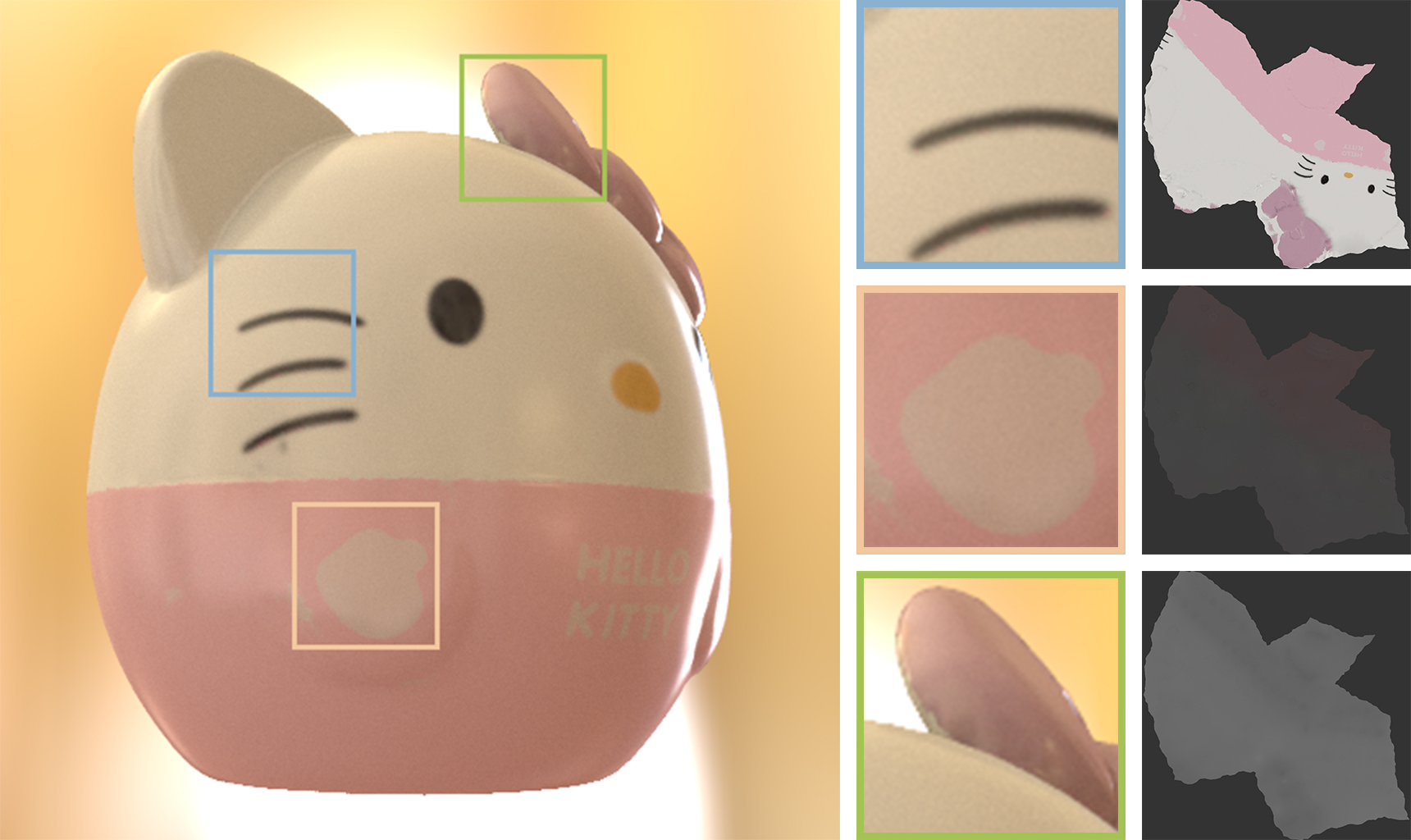} &
		\includegraphics[height=\resLen]{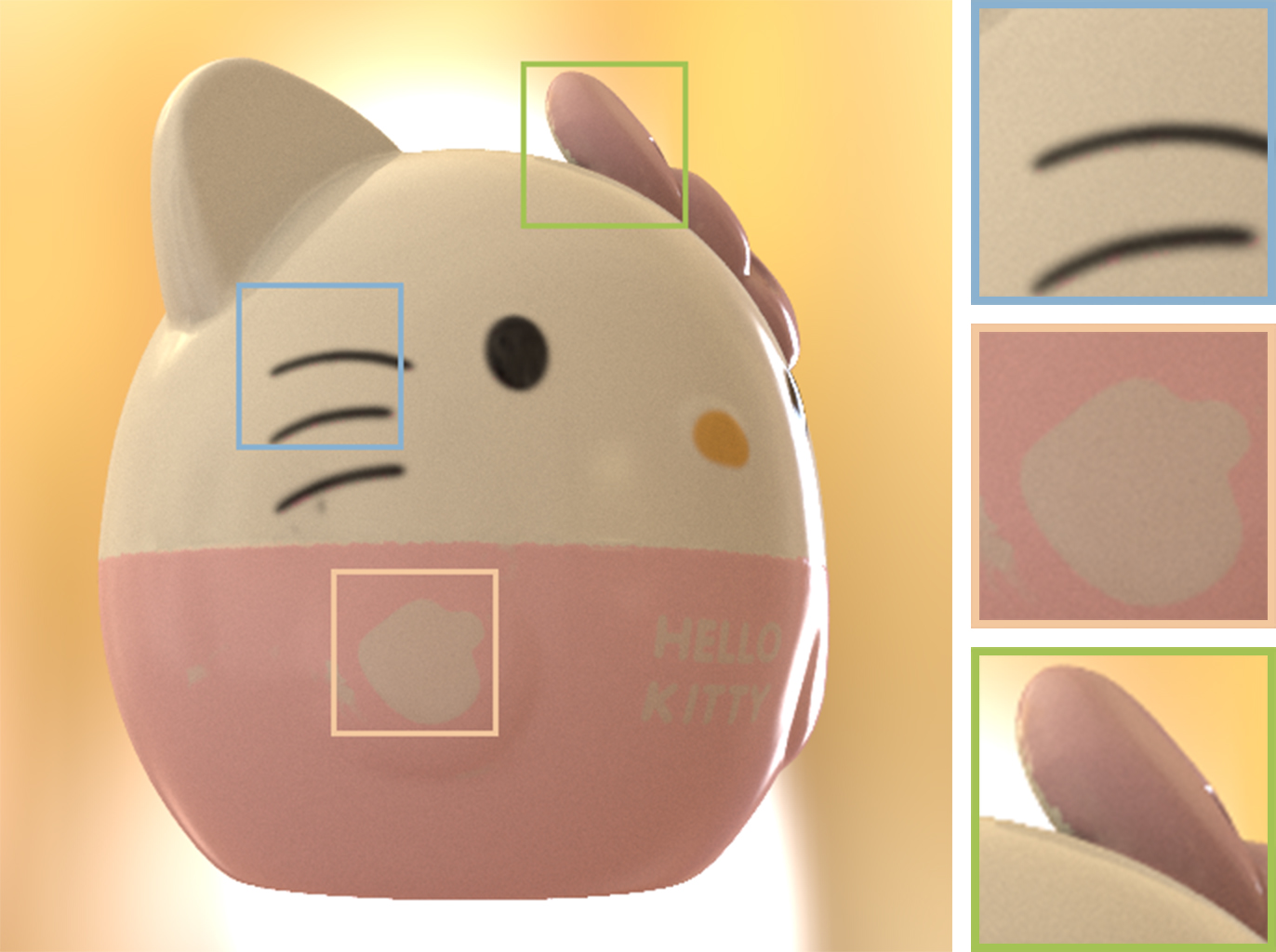}\\
		RMSE: 0.0472 & RMSE: 0.0636 & RMSE: \textbf{0.0177} & \textit{kitty}\\[2pt] 
  	\end{tabular}
    \caption{\label{fig:ablation4}
    	\textbf{Ablation study} on our material loss of Eq.~\protect\eqref{eq:los_reg_mat}.
    	Using identical initial reflectance maps and optimization configurations, models optimized with various components of the material loss neglected are rendered under a novel environmental illumination.
    	On the right of each reconstruction result, we show the optimized reflectance maps (from top to bottom: diffuse albedo, specular albedo, and roughness).
    }
\end{figure*}

\begin{figure*}[t]
	\centering
	\small
	\setlength{\resLen}{1.26in}
	\addtolength{\tabcolsep}{-5pt}
	\begin{tabular}{ccccc}
		\textsc{\scriptsize 10 inputs} &
		\textsc{\scriptsize 30 inputs} &
		\textsc{\scriptsize 50 inputs} &
		\textsc{\scriptsize 100 inputs} &
		\textsc{\scriptsize GT} \\ 
		\includegraphics[trim=70 65 70 75, clip, width=\resLen]{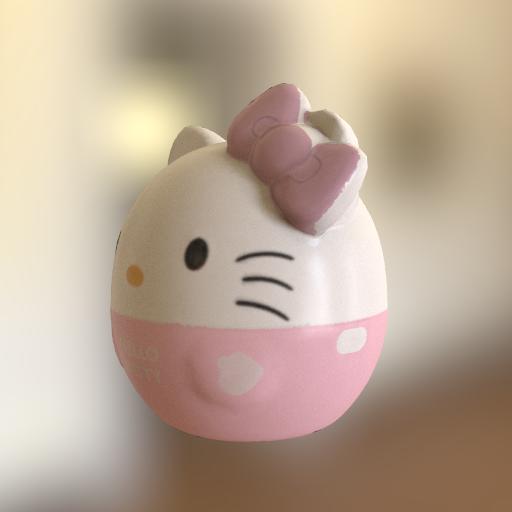} &
		\includegraphics[trim=70 65 70 75, clip, width=\resLen]{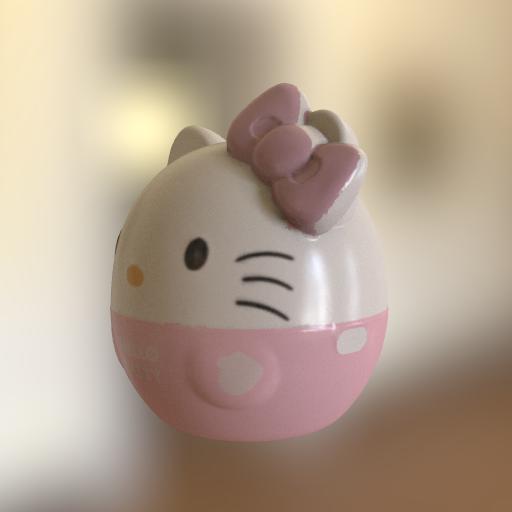} &
		\includegraphics[trim=70 65 70 75, clip, width=\resLen]{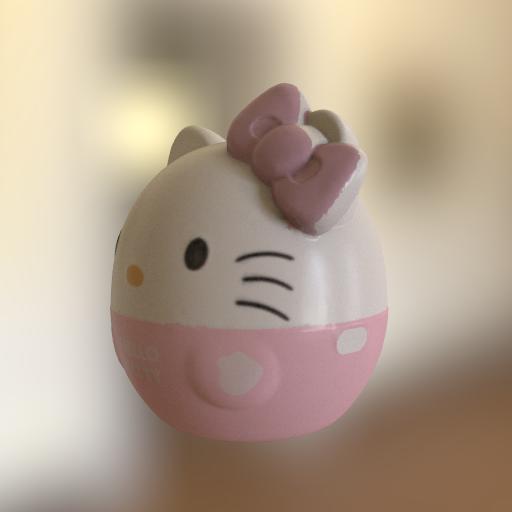} &
		\includegraphics[trim=70 65 70 75, clip, width=\resLen]{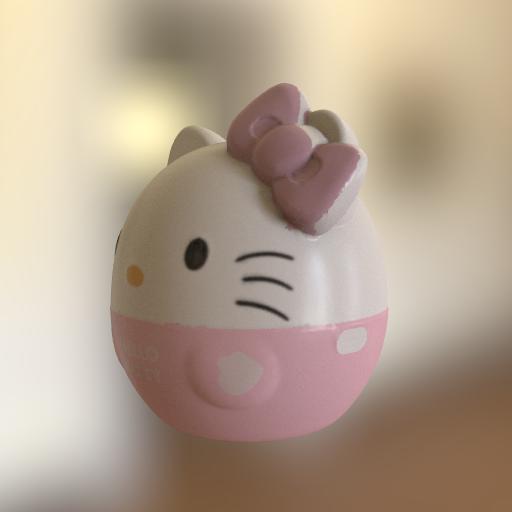} &
		\includegraphics[trim=70 65 70 75, clip, width=\resLen]{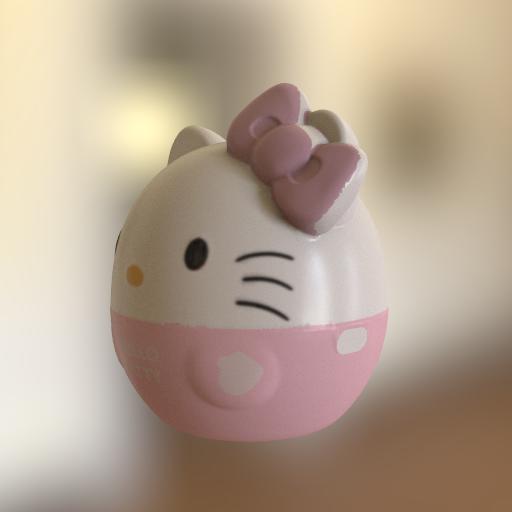}\\
		RMSE: 0.0494 & RMSE: 0.0141 & RMSE: 0.0072 & RMSE: 0.0065 & \textit{kitty}\\[2pt] 
		\includegraphics[trim=73 85 77 65, clip, width=\resLen]{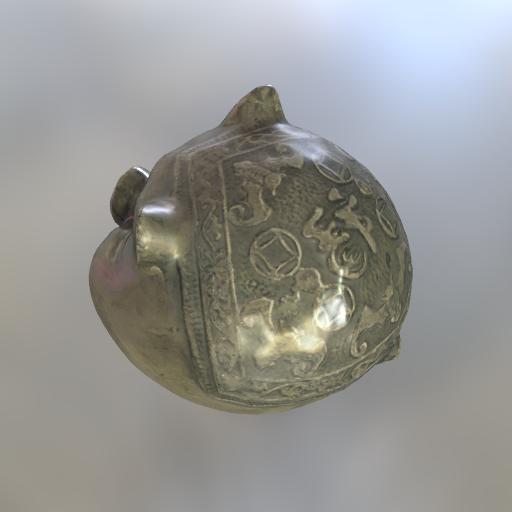} &
		\includegraphics[trim=73 85 77 65, clip, width=\resLen]{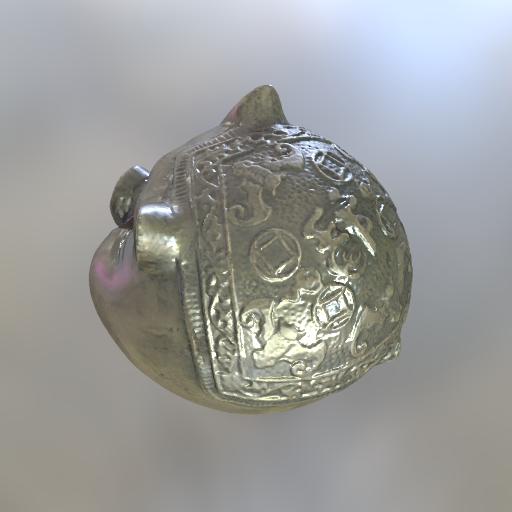} &
		\includegraphics[trim=73 85 77 65, clip, width=\resLen]{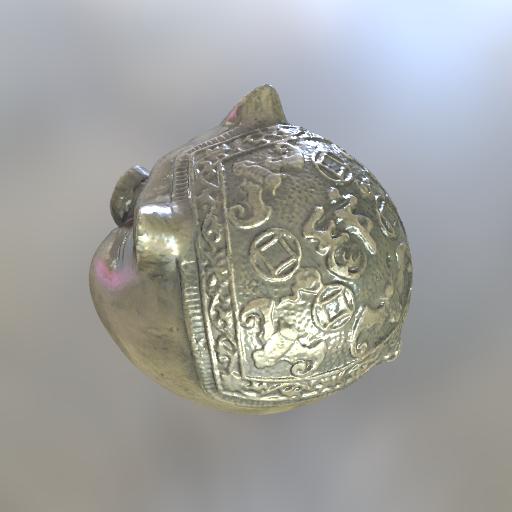} &
		\includegraphics[trim=73 85 77 65, clip, width=\resLen]{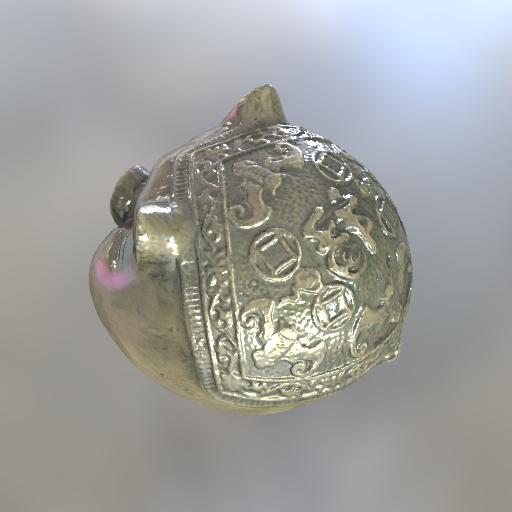} &
		\includegraphics[trim=73 85 77 65, clip, width=\resLen]{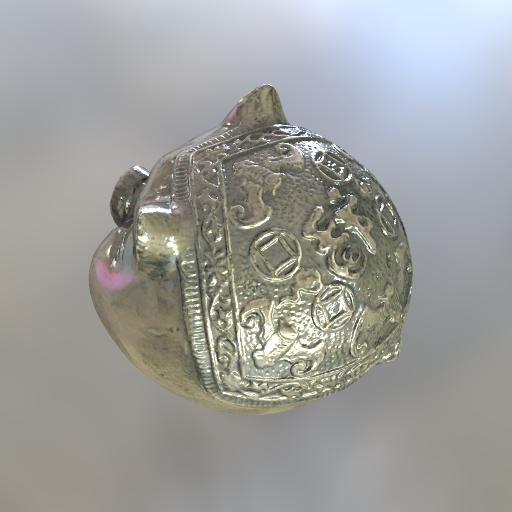} \\
		RMSE: 0.0860 & RMSE: 0.0532 & RMSE: 0.0416 & RMSE: 0.0391 & \textit{pig}\\[2pt] 
	\end{tabular}
	\caption{%
		\small
		\textbf{Ablation study} on number of input images.
		For each object, we show a novel-view rendering of our reconstruction under environmental lighting, given varying number of input images.
		The input images have viewing/lighting positions uniformly sampled around the object.
	}
	\label{fig:ablation_n_input}
\end{figure*}
\begin{figure*}[t]
  \centering
    \setlength{\resLen}{0.94in}
    \newcommand{\ablationTwo}[1]{%
      \includegraphics[trim=60 50 60 70,clip,width=\resLen]{results/ablation2/#1/input_rgb.png} &
      \includegraphics[trim=60 50 60 70,clip,width=\resLen]{results/ablation2/#1/input_depth.png} &
      \includegraphics[width=\resLen]{results/ablation2/#1/colmap3.jpg} &
      \includegraphics[width=\resLen]{results/ablation2/#1/kf3.jpg} &
      \includegraphics[width=\resLen]{results/ablation2/#1/kf_init3.jpg} &
      \includegraphics[width=\resLen]{results/ablation2/#1/ours3.jpg} &
      \includegraphics[width=\resLen]{results/ablation2/#1/gt2.jpg}%
    }
    \addtolength{\tabcolsep}{-4pt}
    \small
    \begin{tabular}{ccccccc}
      \textsc{\scriptsize (a) Input RGB img.} &
      \textsc{\scriptsize (b) Input GT depth} &
      \textsc{\scriptsize (c) Colmap} &
      \textsc{\scriptsize (d1) KF (high)} &
      \textsc{\scriptsize (d2) KF (low)} &
      \textsc{\scriptsize (e) Ours} &
      \textsc{\scriptsize (f) Ground truth}
      \\
      \ablationTwo{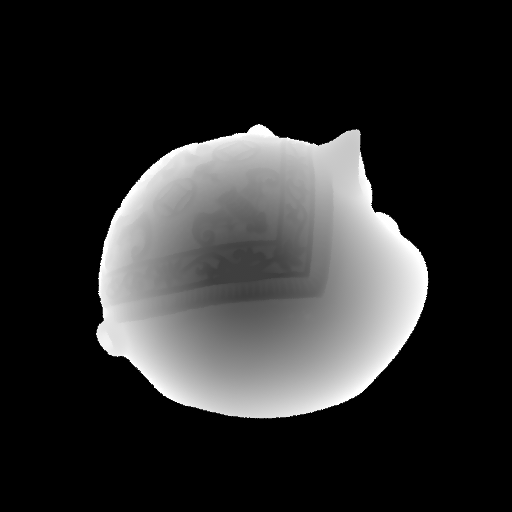}\\[-5pt]
      && 0.0017 & 0.0149 & 0.0318 & \textbf{0.0006} & \textit{pig}\\
      \ablationTwo{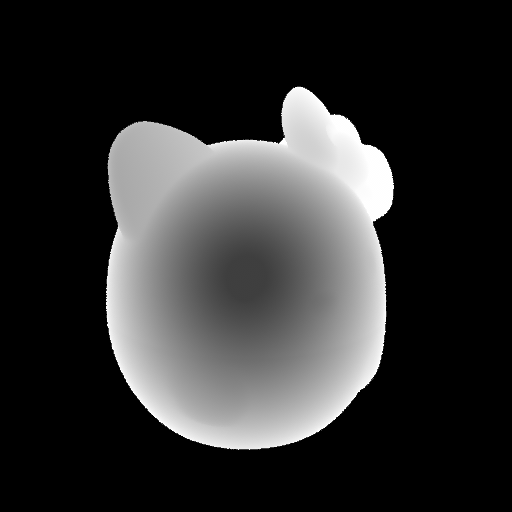}\\[-2pt]
      && 0.0305 & 0.0192 & 0.0324 & \textbf{0.0010} & \textit{kitty}\\[2pt]
    \end{tabular}
    \caption{\label{fig:ablation2}
      \textbf{Comparison} with COLMAP~\cite{schoenberger2016mvs} and Kinect Fusion~\cite{newcombe2011kinectfusion} using synthetic inputs.
      The \emph{COLMAP} results (c) are generated using 50 RGB images (a) with exact camera poses; the \emph{KF-High} (d1) and \emph{KF-Low} (d2) results are created using 50 ground-truth depth images (b) and low-resolution noisy ones, respectively.
      Our method (e), when initialized with \emph{KF-Low} (d2) and using RGB inputs (a), produces much more accurate geometries than \emph{COLMAP} and \emph{KF-High}.
      Similar to Figure~\protect\ref{fig:ablation1}, the numbers indicate average point-to-mesh distances.
    }
\end{figure*}

\subsection{Analysis-by-synthesis optimization}
\label{ssec:opt}
We now present our analysis-by-synthesis optimization pipeline that minimizes Eq.~\eqref{eq:inv_render}.

\paragraph*{Object parameters.}
As stated in \S\ref{ssec:pbdr}, we depict object geometry using a triangle mesh (which is comprised of per-vertex positions $\vtx$ and UV coordinates $\uv$ as well as per-triangle vertex indices) and reflectance using three 2D texture maps specifying the object's spatially varying diffuse albedo~$\albedoD$, specular albedo~$\albedoS$, and surface roughness~$\roughness$, respectively.
In this way, our combined geometry and reflectance parameters are given by $\params = (\vtx, \uv, \albedoD, \albedoS, \roughness)$.
Note, we do not modify the connectivity of the triangle vertices and rely on additional re-meshing steps, which we will discuss in \S\ref{ssec:opt_c2f}, to improve mesh topology.

\paragraph*{Loss.}
A key ingredient in our analysis-by-synthesis optimization is the loss $\Ltot$.
Let $\Itargets := (\Itarget_1, \Itarget_2, \ldots)$ be a set of images of some object (with camera location and pose calibrated for each image $\Itarget_k$).
Then, our loss takes the form:
\begin{equation}
	\label{eq:los1}
	\Ltot(\Irenders(\params), \params;\ \Itargets) :=
	\Lrender(\Irenders(\params);\ \Itargets) + \Lreg(\params),
\end{equation}
where $\Lrender$ is the \emph{rendering} loss that measures the difference between rendered and target object appearances.
Specifically, we set
\begin{equation}
	\label{eq:los_render}
	\Lrender(\Irenders(\params);\ \Itargets) := \wRender \textstyle\sum_k \left\| \Phi_k(\Irender_k(\params)) -  \Phi_k(\Itarget_k) \right\|_1,
\end{equation}
where $\wRender > 0$ is a user-specified weight, $\Irenders(\params) := (\Irender_1(\params), \Irender_2(\params), \ldots)$ denotes images rendered using our forward-rendering model of Eq.~\eqref{eq:render} with object geometry and reflectance specified by $\params$ (under identical camera configurations as the input images), and $\Phi_k$ captures pixel-wise post-processing operations such as tone-mapping and background-removing masking.

We estimate gradients of the rendering loss of Eq.~\eqref{eq:los_render} with respect to the object parameters $\params$ using our differentiable rendering method described in \S\ref{ssec:pbdr}.
We will demonstrate in \S\ref{ssec:eval} that accurate gradients 
are crucial to obtain high-quality optimization results.

In Eq.~\eqref{eq:los1}, $\Lreg(\params)$ is a \emph{regularization} term for improving the robustness of the optimization, which we will discuss in \S\ref{ssec:opt_reg}.
Gradients of this term can be obtained easily using automatic differentiation.

\paragraph*{Optimization process.}
Like any other analysis-by-synthesis method, our technique takes as input an initial configuration of an object's geometry and reflectance.
In practice, we initialize object geometry using MVS or Kinect Fusion.
Our technique is capable of producing high-quality reconstructions using crude initializations (obtained using low-resolution and noisy inputs).
For the reflectance maps, we simply initialize them as constant-valued textures.

Provided an initial configuration of the object's geometry and reflectance, we minimize the loss of Eq.~\eqref{eq:los1} using the Adam algorithm~\cite{kingma2014adam}.

Further, to make the optimization more robust, we leverage a coarse-to-fine approach that periodically performs remeshing and upsamples the reflectance-describing textures.
We will provide more details on this process in \S\ref{ssec:opt_c2f}.

\subsection{Regularization}
\label{ssec:opt_reg}
Using only the rendering loss $\Lrender$ expressed in
Eq.~\eqref{eq:los_render} can make the optimization unstable and/or
converge to local minima.
To address this problem, we regularize the optimization by introducing another loss $\Lreg$ that in turn consists of a \emph{material} loss~$\Lmat$ and a \emph{mesh} one $\Lmesh$:
\begin{equation}
	\label{eq:los_reg}
	\Lreg(\params) := \Lmesh(\mesh) + \Lmat(\albedoD, \albedoS, \roughness),
\end{equation}
which we will discuss in the following.

\paragraph*{Mesh loss.}
We encourage our optimization to return ``smooth'' object geometry by introducing a \emph{mesh loss}:
\begin{equation}
	\label{eq:los_reg_mesh}
	\Lmesh(\mesh) := \Llap(\mesh) + \Lnorm(\mesh) + \Ledge(\mesh),
\end{equation}
where the \emph{mesh-Laplacian loss}~$\Llap$ of a mesh with $n$ vertices is given by  $\Llap(\mesh) := \wLap \| \bm{L} \bm{V} \|^2$ where $\bm{V}$ is an $n \times 3$ matrix with its $i$-th row storing coordinates of the $i$-th vertex, and $\bm{L} \in \Real^{n \times n}$ is the mesh's Laplacian matrix~\cite{nealen2006laplacian}.
%

Additionally, we use a \emph{normal-consistency loss}~$\Lnorm$ to encourage normals of adjacent faces to vary slowly by setting
$\Lnorm(\mesh) := \wNorm \sum_{i, j} [1 - (\norm_i \cdot \norm_j)]^2$, where the sum is over all pairs $(i, j)$ such that the $i$-th and the $j$-th triangles share a common edge, and $\norm_i$ and $\norm_j$ denote the normals of these triangles. 

Lastly, we penalize the mesh for having long edges, which usually yield ill-shaped triangles, by utilizing an \emph{edge-length loss}
$\Ledge := \wEdge\, (\sum_i e_i^2)^{\nicefrac{1}{2}}$, where $e_i$ denotes the length of the $i$-th face edge. 

\paragraph*{Material loss.}
Our material loss $\Lmat$ regularizes the reflectance maps representing diffuse albedo $\albedoD$, specular albedo $\albedoS$, and surface roughness $\roughness$.
Specifically, we set
\begin{equation}
	\label{eq:los_reg_mat}
	\Lmat(\albedoD, \albedoS, \roughness) := \Lspec(\albedoD, \albedoS) + \Lroug(\roughness),
\end{equation}
where $\Lspec$ correlates diffuse and specular albedos~\cite{schmitt2020joint}: assuming nearby pixels with similar diffuse albedos to have similar specular ones, we set
$\Lspec(\albedoS, \albedoD) :=  \wSpec \sum_{\bp} \,\| \albedoS[\bp] - (\sum_{\bq} \albedoS[\bq] \, \mu_{\bp,\bq})/(\sum_{\bq} \mu_{\bp,\bq}) \|_1$,
where $\mu_{\bp,\bq} := \exp\big(-\frac{\|\bp - \bq\|_2^2}{2\sigma_1^2} - \frac{(\albedoD[\bp] - \albedoD[\bq])^2}{2\sigma_2^2}\big)$ is the bilateral weight between pixels with indices $\bp, \bq \in \mathbb{Z}^2$.


Spatially varying surface roughness is known to be challenging to optimize even when the object geometry is known~\cite{gao2019deep}.
To regularize our optimization of surface roughness, we introduce a smoothness term that measures its total variation: $\Lroug(\roughness) := \wRoug \sum_{i, j} ( |\roughness[i+1, j] - \roughness[i, j] | + |\roughness[i, j+1] - \roughness[i, j]| )$, where $\roughness[i, j]$ indicate the value of the $(i, j)$-th pixel in the roughness map.


%
\subsection{Improving robustness}
\label{ssec:opt_c2f}
As described in \S\ref{ssec:opt}, when minimizing the loss of Eq.~\eqref{eq:los1}, we keep the mesh topology unchanged.
This, unfortunately, can severely limit the flexibility of our optimization of object geometry, making the result highly sensitive to the quality of the initial mesh.
Additionally, without taking precautions, updating vertex positions can introduce artifacts (e.g., self intersections) to the mesh that cannot be easily fixed by later iterations.

To address these problems, we utilize a few extra steps. 

\paragraph*{Coarse-to-fine optimization.}
Instead of performing the entire optimization at a single resolution, we utilize a coarse-to-fine process for improved robustness.
Similar steps have been taken in several prior works, although typically limited to either geometry~\cite{sharf2006competing, sahilliouglu2010coarse, kazhdan2013screened, tsai2019beyond} or reflectance~\cite{dong2014appearance, riviere2016mobile, hui2017reflectance}.

Specifically, we start the optimization by using low-resolution meshes and reflectance maps.
If the input already has high resolutions, we simplify them via remeshing and image downsampling.
Our optimization process then involves multiple stages.
During each stage, we iteratively refine the object geometry and reflectance with fixed mesh topology.
After each stage (except the final one), we upsample the mesh (using 
instant meshes~\cite{Jakob2015Instant}) and the texture maps (using simple bilinear interpolation).


\paragraph*{Robust surface evolution.}
During optimization, if the vertex positions are updated na\"ively (i.e., using simple gradient-based updates with no validation checks), the mesh can have degraded quality and even become non-manifold (i.e., with artifacts like holes and self-intersections).
Motivated by other optimization-driven mesh editing algorithms~\cite{sacht2015nested, liu2018paparazzi}, we evolve a mesh using a pipeline implemented in the El Topo library~\cite{elTopo}: Given the initial positions of a set of vertices with associated displacements, El Topo moves each vertex along its displacement vector while ensuring no self-intersection is generated.

 \begin{figure*}[t]
	\centering
 	\includegraphics[width=\textwidth]{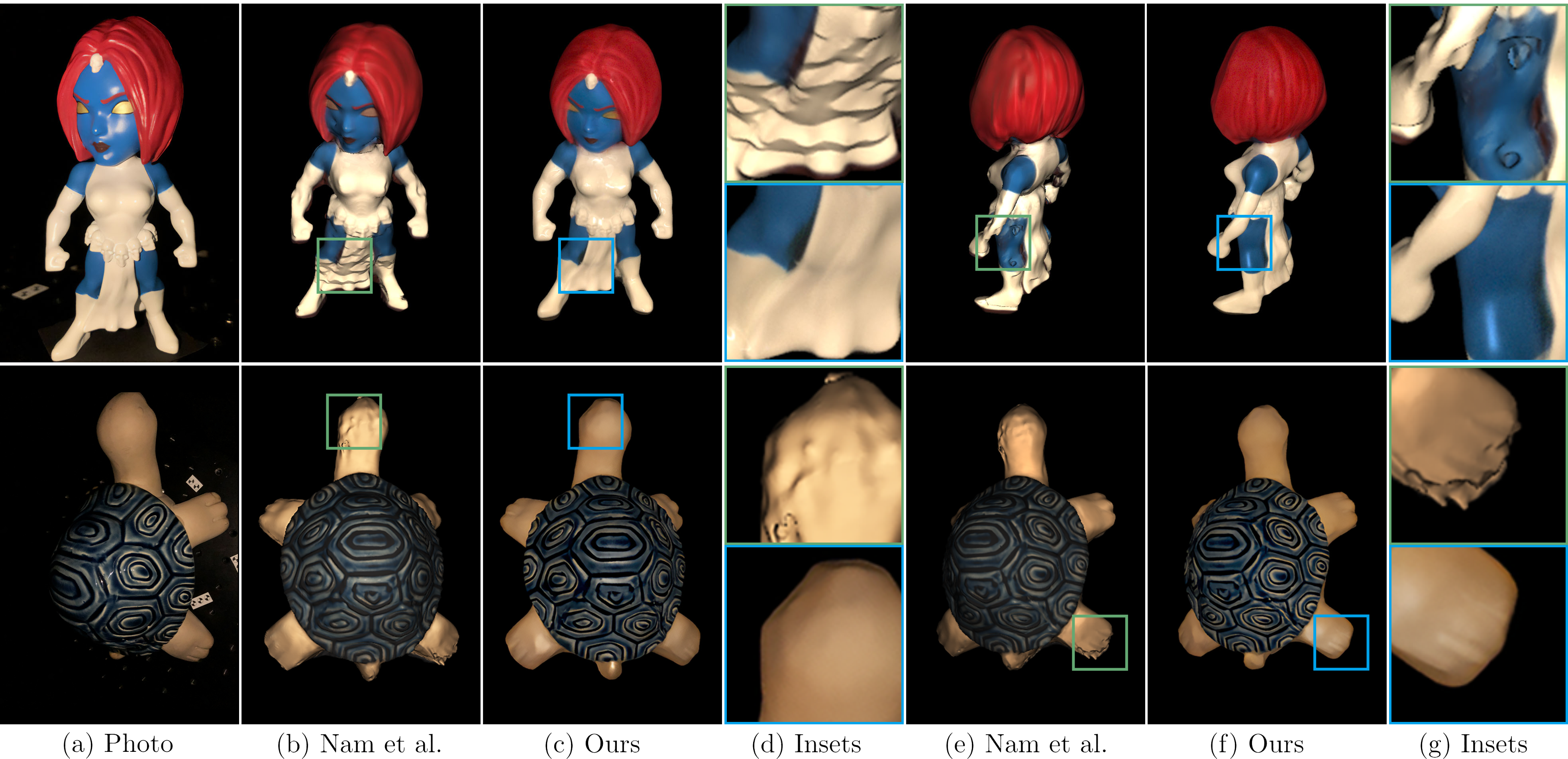}
	\caption{\label{fig:nam}
		\textbf{Comparison} with Nam et al.~\cite{nam2018practical}: We render the reconstructed object under novel view. Nam et al.'s method produces bumpy geometry and inaccurate highlights. In contrast, our method 
		produces much cleaner results that closely resemble the real object.
	}
\end{figure*}

\begin{table}[t]
	\centering
	\small
	\caption{\label{tab:renderer_perf}
		\textbf{Rendering performance}. We report the rendering cost in seconds (averaged across 100 times) of each differentiable renderer in resolution $512 \times 512$ and $4$ samples per pixel on a Titan RTX graphics card. 
	}
	\vspace{1em}
	\addtolength{\tabcolsep}{-3pt}
	\begin{tabular}{lcccccc}
		\toprule
		       &  SoftRas & PyTorch3D & Mitsuba 2 & Nvdiffrast    & Redner & Ours  \\ 
		Kettle &  0.0184  & 0.0202    & 0.0877    & 0.0013        & 0.1196 & 0.0143\\
		Maneki &  0.0192  & 0.0224    & 0.0863    & 0.0010        & 0.1029 & 0.0146\\
		Pig    &  0.0971  & 0.0772    & 0.0913    & 0.0014        & 0.1336 & 0.0263\\
		Kitty  &  0.0249  & 0.0334    & 0.0889    & 0.0010        & 0.1190 & 0.0225\\
		\bottomrule
	\end{tabular}%
\end{table}


\section{Results}
\label{sec:results}
%
%
We implement our differentiable renderer (\S\ref{ssec:pbdr}) in C++ with CUDA 11 and OptiX 7.1.
For vectorized and differentiable computations, we utilize the Enoki library~\cite{Enoki}, which has been demonstrated to be more efficient than generic ones like Tensorflow and PyTorch for rendering~\cite{nimier2019mitsuba}.

We implement the rest of our optimization pipeline, including the loss computations, in PyTorch.
We use one set of weights for our optimizations: $\wRender = 1$ for the rendering loss of Eq.~\eqref{eq:los_render}; $(\wLap, \wEdge, \wNorm) = (0.1, 1, 0.01)$ for the mesh loss of Eq.~\eqref{eq:los_reg_mesh} and $(\wSpec, \wRoug) = (0.01, 0.001)$ for the material loss of Eq.~\eqref{eq:los_reg_mat}.

In practice, our optimization involves 500--1000 iterations (for all coarse-to-fine stages) and takes 0.5--2 hours per example (see the supplement for more details).

\subsection{Evaluations and comparisons}
\label{ssec:eval}
\noindent \emph{Please see the supplemental material for more results.}

\paragraph*{Comparison with differentiable renderers.}
Our renderer enjoys high performance, thanks to its specialized nature (i.e., focused on the collocated configuration) and the combined efficiency of RTX ray tracing offered by OptiX and GPU-based differentiable computations by Enoki.
As demonstrated in Table~\ref{tab:renderer_perf}, our renderer is faster than SoftRas~\cite{liu2019soft}, PyTorch3D~\cite{ravi2020accelerating}, and Mitsuba~2~\cite{nimier2019mitsuba} without the need to introduce bias to the gradient estimates.
Nvdiffrast~\cite{Laine2020diffrast} is faster than our system but produces approximated gradients.
Lastly, compared to Redner~\cite{li2018differentiable}, another differentiable renderer that produces unbiased gradients, our renderer offers better performance.

To further demonstrate the importance for having accurate gradients, we conduct a synthetic experiment where the shape of an object is optimized (with known diffuse reflectance).
Using identical input images, initial configurations, losses, and optimization settings (e.g., learning rate), we ran multiple optimizations using Adam~\cite{kingma2014adam} with gradients produced by SoftRas, PyTorch3D, Mitsuba~2, Nvdiffrast, and our technique, respectively.
As shown in Figure~\ref{fig:ablation1}, using biased gradients yields various artifacts or blurry geometries in the optimized results.

\begin{figure*}[t]
 	\centering
   	\setlength{\resLen}{0.95in}
   	\newcommand{\resultsSyn}[1]{%
   		\begin{overpic}[width=\resLen]{results/synthetic/#1/KF2.jpg}
   			\put(2, 88) {\small\textcolor{black}{\bfseries #1}}
   		\end{overpic}
   		&
		\includegraphics[width=\resLen]{results/synthetic/#1/novel_tar2.jpg} &
		\includegraphics[width=\resLen]{results/synthetic/#1/novel_fit2.jpg} &
		\includegraphics[width=\resLen]{results/synthetic/#1/envmap1_tar2.jpg} &
		\includegraphics[width=\resLen]{results/synthetic/#1/envmap1_fit2.jpg} &
		\includegraphics[width=\resLen]{results/synthetic/#1/envmap2_tar2.jpg} &
		\includegraphics[width=\resLen]{results/synthetic/#1/envmap2_fit2.jpg}%
	}
   	\addtolength{\tabcolsep}{-5pt}
	\begin{tabular}{ccccccc}
		\textsc{\scriptsize Init mesh} &
		\textsc{\scriptsize GT (point-light)} &
		\textsc{\scriptsize Ours (point-light)} &
		\textsc{\scriptsize GT (env. map 1)} &
		\textsc{\scriptsize Ours (env. map 1)} &
		\textsc{\scriptsize GT (env. map 2)} &
		\textsc{\scriptsize Ours (env. map 2)}\\
		\resultsSyn{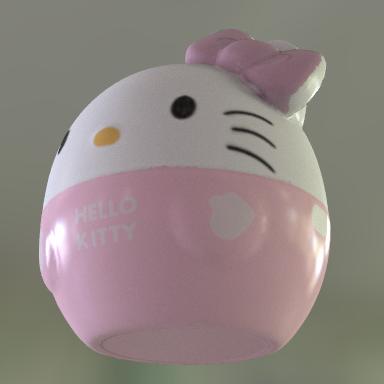}\\[-2pt]
		\resultsSyn{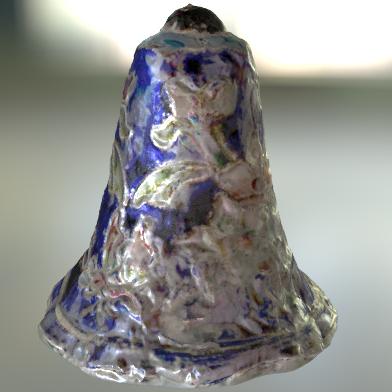}\\[-2pt]
		\resultsSyn{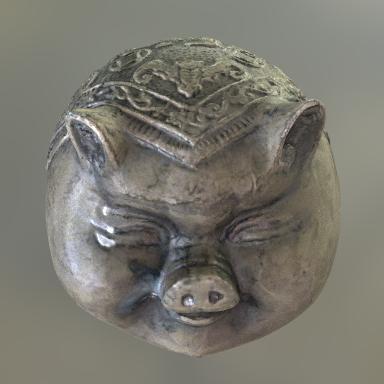}
	\end{tabular}
    \caption{\label{fig:main}
    	\textbf{Reconstruction results} using synthetic inputs:
    	We obtain the initial meshes (in the left column) using Kinect Fusion~\cite{newcombe2011kinectfusion} with low-resolution ($48 \times 48$) and noisy depths.
    	Our analysis-by-synthesis pipeline successfully recovers both geometric and reflectance details, producing high-fidelity results under novel viewing and illumination conditions.
    } 
   
\end{figure*}

\paragraph*{Effectiveness of shape optimization.}
To ensure robustness when optimizing the shape of an object, our technique utilizes a mesh loss (\S\ref{ssec:opt_reg}) as well as a coarse-to-fine framework (\S\ref{ssec:opt_c2f}).
We conduct another experiment to evaluate the effectiveness of these steps.
Specifically, we optimize the shape of the \emph{pig} model using identical optimization configurations except for (i)~having various components of the mesh loss turned off; and (ii) not using the coarse-to-fine framework.
As shown in Figure~\ref{fig:ablation3}, with the mesh Laplacian
loss~$\Llap$ neglected (by setting $\wLap = 0$), the resulting
geometry becomes ``bumpy''; without the normal and edge-length losses
$\Lnorm$ and $\Ledge$, the optimized geometry also has artifacts due
to sharp normal changes and ill-shaped triangles. Additionally,
without performing the optimization in a coarse-to-fine fashion (by
directly starting with a subdivided version of the initial mesh), the
optimization gets stuck in a local optimum (with all losses enabled).

\paragraph*{Effectiveness of material loss.}
Our material loss of Eq.~\eqref{eq:los_reg_mat} is important for obtaining clean reflectance maps that generalize well to novel settings.
As shown in Figure~\ref{fig:ablation4}, without correlating diffuse and specular albedos (by having $\wSpec = 0$), diffuse colors are ``baked'' into specular albedo, leading to heavy artifacts under novel environmental illumination.
With the roughness smoothness~$\Lroug$ disabled, the resulting roughness map contains high-frequency noise that leads to artifacts in rendered specular highlights.

\paragraph*{Number of input images.}
We evaluate how the number of input images affects the reconstruction quality of our method in Figure~\ref{fig:ablation_n_input}.
Using a small number (e.g., 10) of images, the optimization becomes highly under-constrained, making it difficult for our model to produce accurate appearance under novel viewing conditions.
The accuracy our novel-view renderings improves quickly as the number of input images increases: With 50 or more input images, our renderings closely match the groundtruth.

\paragraph*{Comparison with previous methods.}
To further evaluate the effectiveness of our technique for recovering object geometry, we compare with several previous methods \cite{schoenberger2016mvs,newcombe2011kinectfusion,nam2018practical}.

Figure~\ref{fig:ablation2} shows comparisons with COLMAP~\cite{schoenberger2016mvs} and Kinect Fusion~\cite{newcombe2011kinectfusion} using synthetic inputs.
Our technique, when using crude initial geometries (obtained using Kinect Fusion with low-resolution and noisy depth images), produces results with much higher quality than the baselines.
COLMAP fails badly for the \emph{kitty} example since the object contains insufficient textures for correspondences to be reliable established.

Additionally, we compare our technique with the work from Nam~et~al.~\cite{nam2018practical} using real inputs (i.e., photographs).
As demonstrated in Figure~\ref{fig:nam}, Nam~et~al.'s method does not explicitly optimizes object geometry based on image losses and returns geometries with heavy artifacts.
Our technique, on the other hand, is much more robust and capable of reproducing the clean geometry and appearance of the physical model.

\subsection{Reconstruction results}
\label{ssec:more_results}
Figure~\ref{fig:main} shows reconstruction results obtained using synthetic input images and rendered under novel views and illuminations.
The initial geometries are obtained using Kinect Fusion with low-resolution noisy depth inputs.
Using $50$ input images, our technique offers the robustness for recovering both smooth (e.g., the \textit{kitty} example) and detailed (e.g., the \textit{pig} example) geometries and reflectance.

\begin{figure*}[t]
 	\centering
	\setlength{\resLen}{0.84in}
    	\newcommand{\resultsReal}[3]{%
		\begin{overpic}[width=\resLen]{results/real/#1/img#2.jpg}
			\put(2, 4) {\small\textcolor{white}{\bfseries #1}}
		\end{overpic}
		&
		\includegraphics[width=\resLen]{results/real/#1/our_#3.jpg} &
		\includegraphics[width=\resLen]{results/real/#1/normal_#3.jpg} &
		\includegraphics[width=\resLen]{results/real/#1/novel.jpg}%
	}
   	\addtolength{\tabcolsep}{-5.4pt}
	\begin{tabular}{cccccccc}
		\textsc{\scriptsize Photo. (novel)} &
		\textsc{\scriptsize Ours} &
		\textsc{\scriptsize Normal} &
		\textsc{\scriptsize Ours (env. map)} &
		\textsc{\scriptsize Photo. (novel)} &
		\textsc{\scriptsize Ours} &
		\textsc{\scriptsize Normal} &
		\textsc{\scriptsize Ours (env. map)}
		\\
		\resultsReal{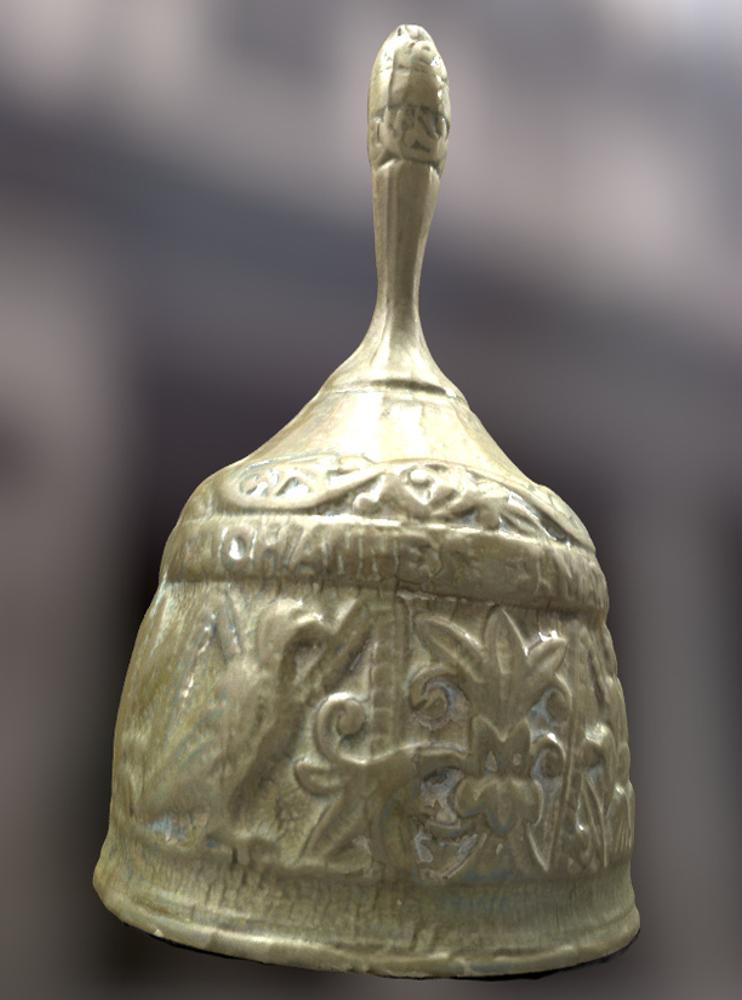}{0229}{229} &
		\resultsReal{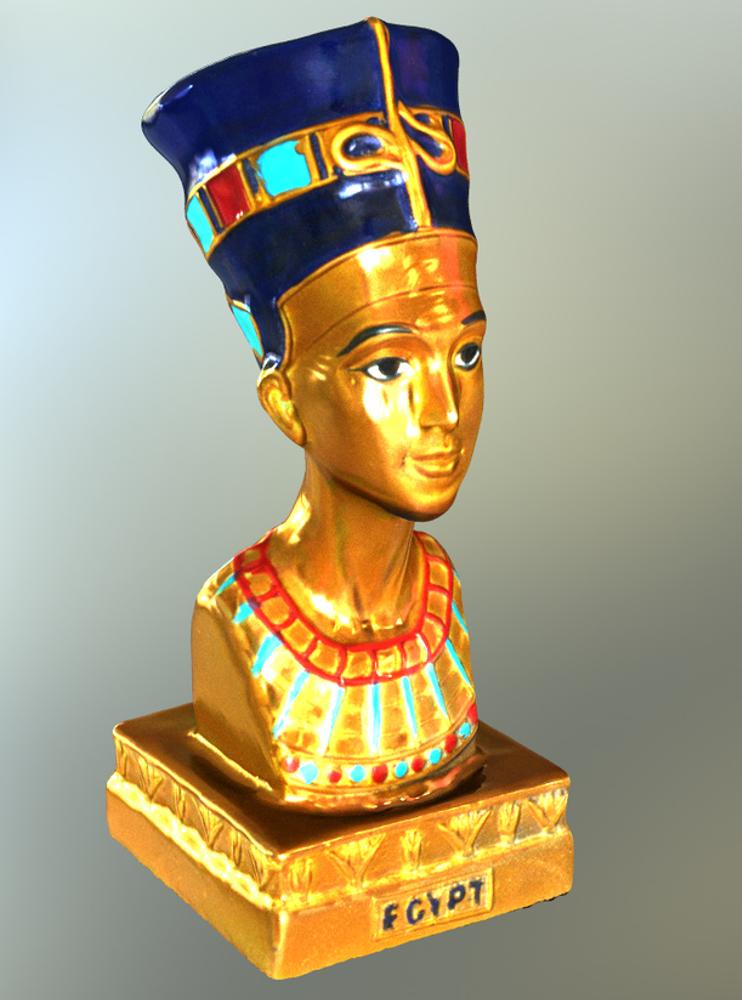}{0001}{1}\\[-2pt]
		\resultsReal{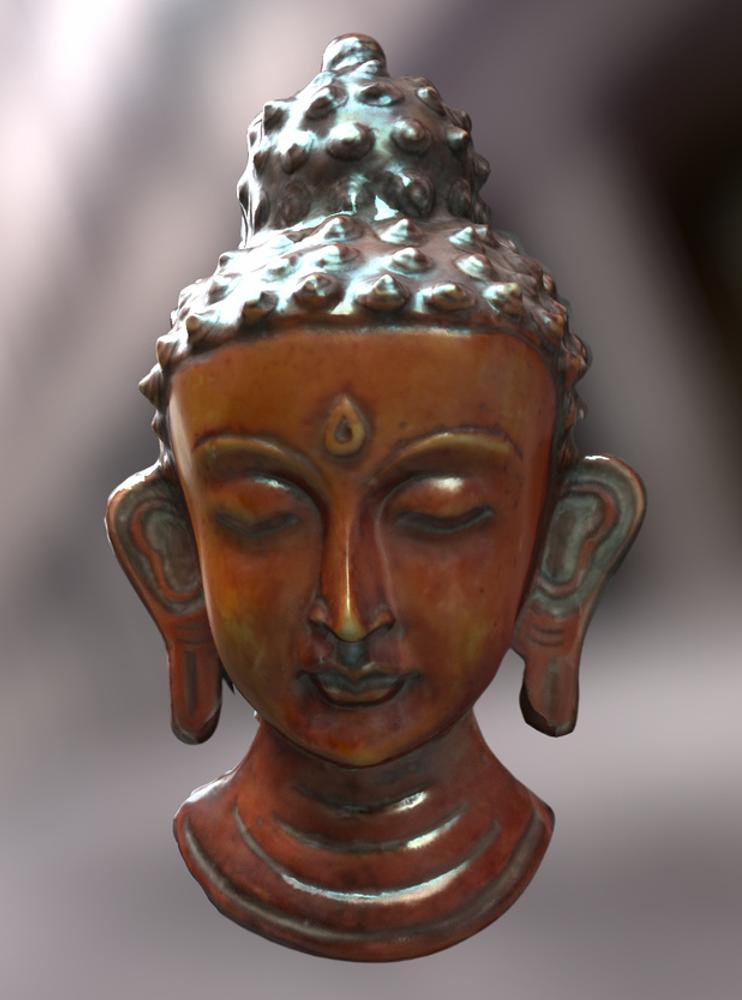}{0072}{72} &
		\resultsReal{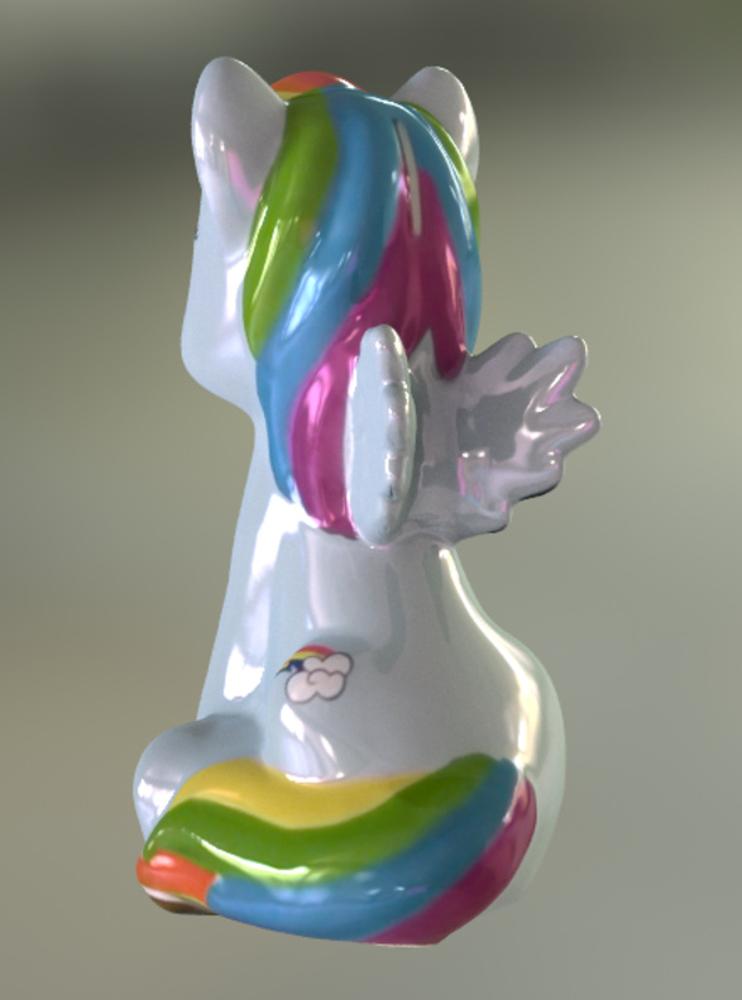}{0165}{165}
	\end{tabular}
    \caption{\label{fig:main2}
    	\textbf{Reconstruction results} of real-world objects:
    	Similar to the synthetic examples in Figure~\protect\ref{fig:main}, our technique recovers detailed object geometries and reflectance details, producing high-quality results under novel viewing and illumination conditions.
    }
\end{figure*}

We show in Figure~\ref{fig:main2} reconstruction results using as input $100$ real photographs per example.
The initial geometries are obtained using COLMAP. 
Our analysis-by-synthesis technique manages to accurately recover the detailed geometry and reflectance of each model.

Please note, in Figures~\ref{fig:main} and \ref{fig:main2}, the detailed geometric structures (e.g., those in the \textit{bell}, \textit{pig}, \textit{chime}, and \textit{buddha} examples) fully emerge from the mesh-based object geometries: no normal or displacement mapping is used.

\begin{figure}[t]
	\centering
	\setlength{\resLen}{0.995\columnwidth}
	\includegraphics[width=\resLen]{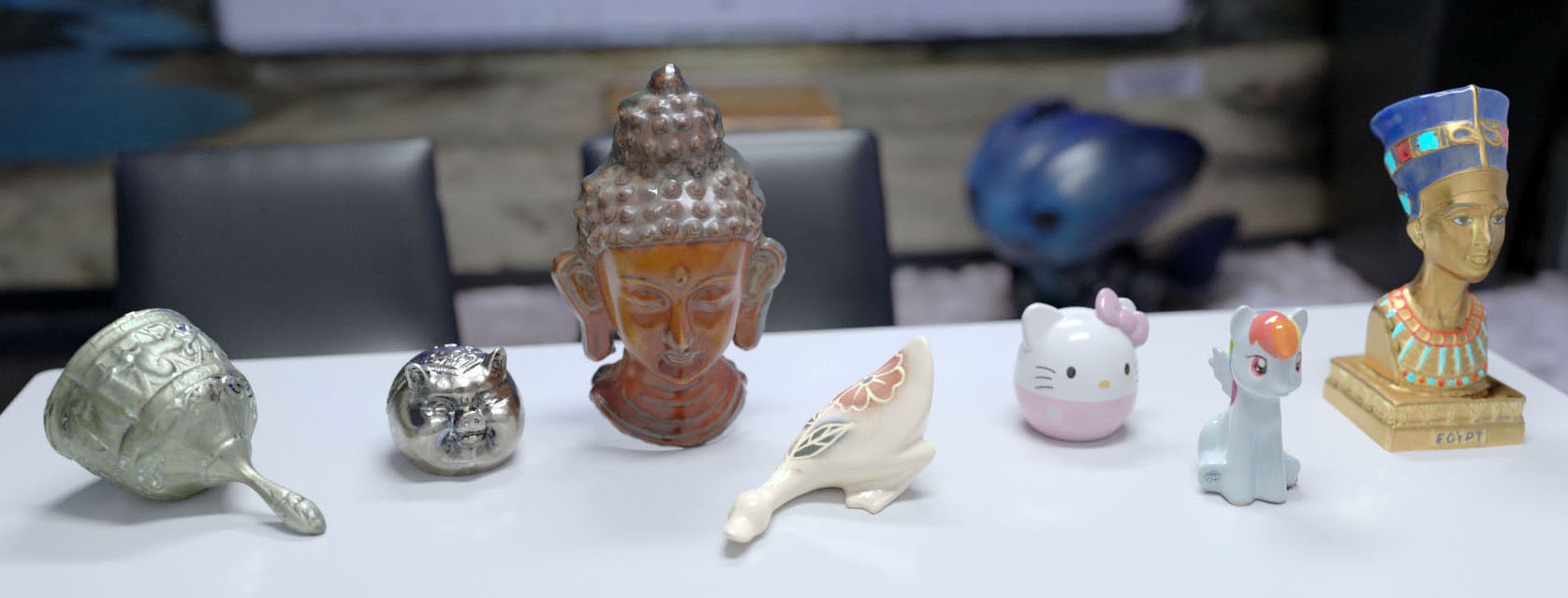}
	\includegraphics[width=\resLen]{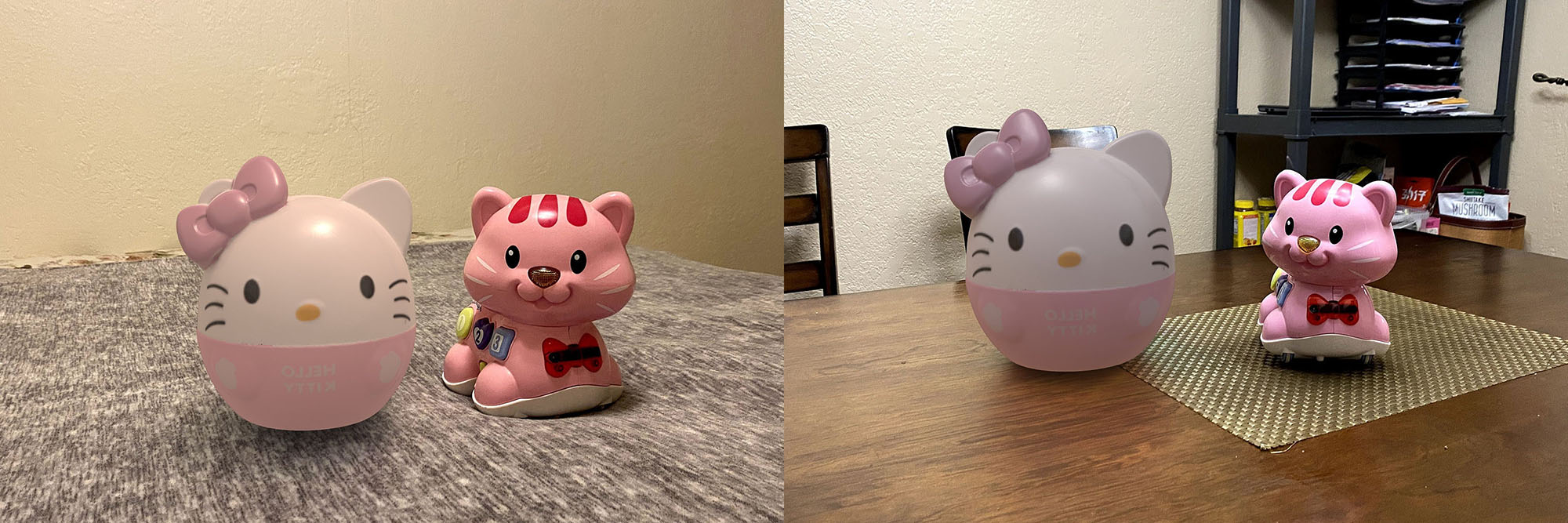}
	\caption{\label{fig:app}
		\textbf{Applications:} The high-quality models generated by our technique can be used for 3D digital modeling (top) and object insertion in augmented reality (bottom).
	}
\end{figure}

Lastly, since our reconstructed models use standard mesh-based representations, they can be used in a broad range of applications (see Figure~\ref{fig:app}).

\revision{
\subsection{Discussion and Analysis}
We believe the quality gain to be obtained for three main reasons: First, we use Monte Carlo edge sampling~\cite{li2018differentiable} that provides accurate gradients with respect to vertex positions, allowing our method to provide more accurate reconstructions of object geometries (cf. Figure~\ref{fig:nam} against~\cite{nam2018practical}); Second, we exploit robust surface evolution, e.g., elTopo, on top of gradient descent, which ensures a manifold mesh (i.e., without self-intersections or other degenerated cases) after every iteration; Third, our coarse-to-fine strategy and the other regularization terms have come together to make our pipeline more robust in practice.

\paragraph*{Failure cases.} Despite our pipeline being robust for most cases in synthetic/real-world settings, failure cases still exist. Firstly, our method has difficulties handling complex changes of the mesh topology---which is a well-known limitation for mesh-based representations. Secondly, when modeling object appearance, our method relies on a simplified version of the Disney BRDF model only dealing with opaque materials, and thus is limited at reconstructing sophisticated surface appearances, such as anisotropic reflection or subsurface scattering.
}



\section{Conclusion}
\label{sec:conclusion}
We introduce a new approach to jointly recover the shape and
reflectance of real-world objects.  At the core of our technique is a
unified analysis-by-synthesis pipeline that iteratively refines object
geometry and reflectance.  Our custom Monte Carlo differentiable
renderer enjoys higher performance than many existing tools (such as
SoftRas, PyTorch3D, and Mitsuba~2). 
More importantly, our renderer produces unbiased geometric gradients that are crucial for obtaining
high-quality reconstructions.  To further improve the robustness of
our optimization, we leverage a coarse-to-fine framework regularized
using a few geometric and reflectance priors.  We conduct several
ablation studies to evaluate the effectiveness of our differentiable
renderer, losses, and optimization strategies.

\paragraph*{Limitations and future work.}
Our technique is specialized to using a collocated camera and point light.
This configuration can have difficulties in capturing materials exhibiting strong retroreflection.
Generalization to other configurations would be useful in the future.

To refine mesh topology, our technique relies on remeshing steps
(between coarse-to-fine stages).  How topology can be optimized in a
robust and flexible fashion is an important problem for future
investigation.

Lastly, more advanced regularizations of geometry and/or appearance may enable high-quality reconstructions with fewer input images.

\paragraph*{Acknowledgements.} We thank Chenglei Wu, Yujia Chen, Christoph lassner, Sai Bi, Zhengqin Li, Giljoo Nam, Yue Dong, Hongzhi Wu, Zhongshi Jiang as well as the anonymous reviewers for their valuable discussions. We thank the digital artist James Warren from Facebook Reality Labs for modeling and rendering the two table scenes, and Inseung Hwang from KAIST for making comparisons with Nam et al.~\cite{nam2018practical}. This work was supported in part by NSF grants 1900783 and 1900927.
\begingroup
	\small
	\bibliographystyle{eg-alpha-doi}
	\bibliography{egbib}
\endgroup
\end{document}


\BibtexOrBiblatex
\electronicVersion

\title{Unified Shape and SVBRDF Recovery using Differentiable Monte Carlo Rendering: Supplemental Material}

\author[Luan et al.]
{\parbox{\textwidth}{\centering 
			Fujun Luan$^{1,3}$,
         	Shuang Zhao$^{2}$,
         	Kavita Bala$^{1}$,
         	Zhao Dong$^{3}$
        }
        \\
{\parbox{\textwidth}{\centering $^1$Cornell University\\
         $^2$University of California, Irvine\\
         $^3$Facebook Reality Labs\\
       }
}
}
\maketitle

In this document, we provide additional analyses and results in addition to those in the main paper.

Specifically, we include detailed information on the performance of our analysis-by-synthesis optimization in \S\ref{sec:opt_perf}.
Then, \S\ref{sec:ablation} contains additional ablation studies and comparisons with existing methods.
Lastly, in \S\ref{sec:recon}, we show additional reconstruction results (using both synthetic and real data).

\section{Optimization performance}
\label{sec:opt_perf}
We provide a detailed analysis on the performance of our technique in
this section.
The table below shows the total optimization time for each example as well as the percentages for differentiable rendering (DR), loss computation (Loss), backpropagation (BP), and geometric processing (GP) that involves steps such as mesh evolution and remeshing.
The performance numbers are acquired on a workstation equipped with an Intel Xeon E5-2630 v3 processor, 64 GB of RAM, and an Nvidia Titan RTX graphics card.

\iffalse
	\begin{table}[htp]
		\centering
		\small
		\caption{\label{tab:supp_perf}
			\textbf{Performance profiling}. We report the per-iteration profiling results (in seconds) of each component in our optimzation framework for synthetic data tested on a Titan RTX graphics card. 
		}
		\vspace{1em}
		\addtolength{\tabcolsep}{-3.0pt}
		\begin{tabular}{lccccc} 
			\toprule
			       &  Forward & Backward  & El Topo    & Geom. Loss & Mat. Loss  \\ 
			Kitty  &  0.0225  & 0.0797    & 0.7589     & 0.1061     & 0.1449     \\
			Duck   &  0.0232  & 0.0996    & 0.7473     & 0.1052     & 0.1425     \\
			Bear   &  0.0231  & 0.0976    & 0.8098     & 0.1083     & 0.1394     \\
			Bell   &  0.0249  & 0.0934    & 0.8632     & 0.1230     & 0.1225     \\
			Pig    &  0.0263  & 0.0955    & 0.8725     & 0.1268     & 0.1382     \\
			\midrule
			Pony       &  0.0249  & 0.0812    & 0.7741     & 0.1051     & 0.1305 \\
			Camera     &  0.0539  & 0.1434    & 1.0863     & 0.5168     & 0.1446 \\
			Chime      &  0.0524  & 0.1283    & 0.9877     & 0.3452     & 0.1343 \\
			Nefertiti  &  0.0558  & 0.1347    & 1.0913     & 0.4617     & 0.1278 \\
			Buddha     &  0.0563  & 0.1384    & 1.1808     & 0.4823     & 0.1532 \\
			\bottomrule
		\end{tabular}%
	\end{table}
\else
	\begin{center}
		\small
		\addtolength{\tabcolsep}{-1pt}
		\begin{tabular}{lrcccc} 
			\toprule
			&  \textbf{Opt. time} & \textbf{DR} & \textbf{Loss} & \textbf{BP} & \textbf{GP} \\ 
			Kitty  		&  18.54 min & 2.02\%  & 22.57\%  & 7.16\%  & 68.25\%  \\
			Duck   		&  18.63 min & 2.07\%  & 22.16\%  & 8.91\%  & 66.86\%  \\
			Bear   		&  19.64 min & 1.96\%  & 21.02\%  & 8.28\%  & 68.74\%  \\
			Bell   		&  40.90 min & 2.03\%  & 20.01\%  & 7.61\%  & 70.35\%  \\
			Pig    		&  41.98 min & 2.09\%  & 21.04\%  & 7.58\%  & 69.29\%  \\
			\midrule
			Pony     	&  27.89 min & 2.23\%  & 21.11\%  & 7.27\%  & 69.39\%  \\
			Camera   	&  64.83 min & 2.77\%  & 34.00\%  & 7.37\%  & 55.86\%  \\
			Chime    	&  82.39 min & 3.18\%  & 29.10\%  & 7.79\%  & 59.93\%  \\
			Nefertiti   &  93.57 min & 2.98\%  & 31.50\%  & 7.20\%  & 58.32\%  \\
			Buddha    	&  100.55 min & 2.80\%  & 31.60\%  & 6.88\%  & 58.72\%  \\
			\bottomrule
		\end{tabular}%
	\end{center}
\fi

We note that the geometric processing (GP) step takes a large fraction of total optimization time.
This is mainly because the elTopo library~\cite{elTopo} is CPU-based and single-threaded.
We expect a better implementation of this library to significantly improve the performance of our technique.

\section{Ablation studies and comparisons}
\label{sec:ablation}
\paragraph*{Additional ablation studies.}
Our technique is capable of generating plausible results using only 10 inputs.
However, with too few input images, our analysis-by-synthesis optimization could become highly under-constrained, causing the reconstruction results to have overly smooth geometries.
In practice, with 50 or more input images, our method can generate high-quality reconstructions that generalize well to novel conditions.
Thus, we use 50 inputs for the synthetic results and 100 for the real ones in both the paper and the rest of this document.

\begin{figure*}[p]
 	\centering
    \setlength{\resLen}{0.8in}
   	\newcommand{\ablationTwo}[5]{%
	   	\includegraphics[trim=#2 #3 #4 #5, clip, width=\resLen]{results/ablation2/#1/supp_gt2.jpg}&
   		\includegraphics[trim=#2 #3 #4 #5, clip, width=\resLen]{results/ablation2/#1/supp_colmap2.jpg} &
   		\includegraphics[trim=#2 #3 #4 #5, clip, width=\resLen]{results/ablation2/#1/supp_kf2.jpg} &
   		\includegraphics[trim=#2 #3 #4 #5, clip, width=\resLen]{results/ablation2/#1/supp_kf_init2.jpg} &
   		\includegraphics[trim=#2 #3 #4 #5, clip, width=\resLen]{results/ablation2/#1/supp_ours2.jpg}%
   	}
   	\newcommand{\ablationTwoErr}[5]{%
   		\includegraphics[trim=#2 #3 #4 #5, clip, width=\resLen]{results/ablation2/#1/supp_tar2.jpg} &
   		\includegraphics[trim=#2 #3 #4 #5, clip, width=\resLen]{results/ablation2/#1/supp_colmap_err.jpg} &
   		\includegraphics[trim=#2 #3 #4 #5, clip, width=\resLen]{results/ablation2/#1/supp_kf_err.jpg} &
   		\includegraphics[trim=#2 #3 #4 #5, clip, width=\resLen]{results/ablation2/#1/supp_kf_init_err.jpg} &
   		\includegraphics[trim=#2 #3 #4 #5, clip, width=\resLen]{results/ablation2/#1/supp_ours_err.jpg}%
   	}
   	\addtolength{\tabcolsep}{-5pt}
   	\small
   	\begin{tabular}{ccccc}
   		\textsc{\scriptsize Ground truth} &
   		\textsc{\scriptsize Colmap} &
   		\textsc{\scriptsize KF (high)} &
   		\textsc{\scriptsize KF (low)} &
   		\textsc{\scriptsize Ours} 
   		\\
   		\ablationTwo{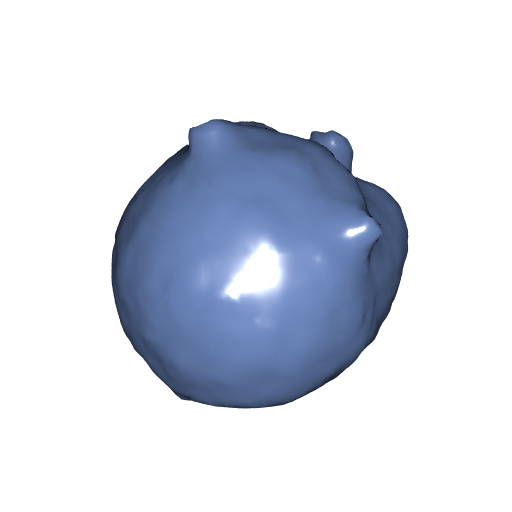}{75}{75}{75}{75}\\[5pt]
   		\ablationTwoErr{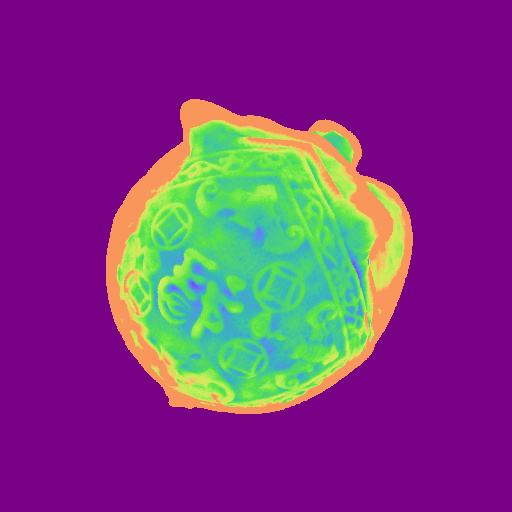}{75}{75}{75}{75}\\
		\textit{pig} & 0.0017 & 0.0149 & 0.0318 & \textbf{0.0006} \\[10pt]
   		\ablationTwo{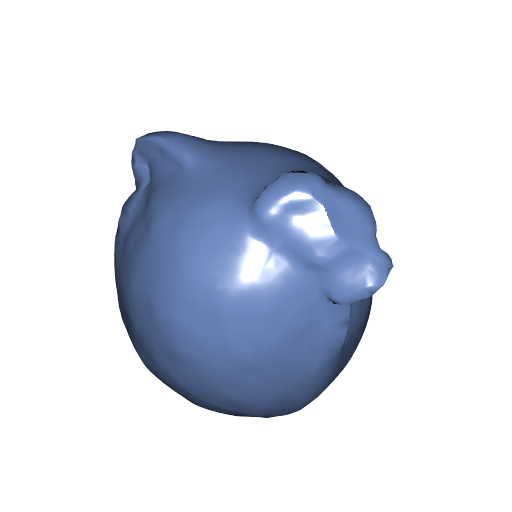}{80}{74}{80}{86}\\[5pt]
   		\ablationTwoErr{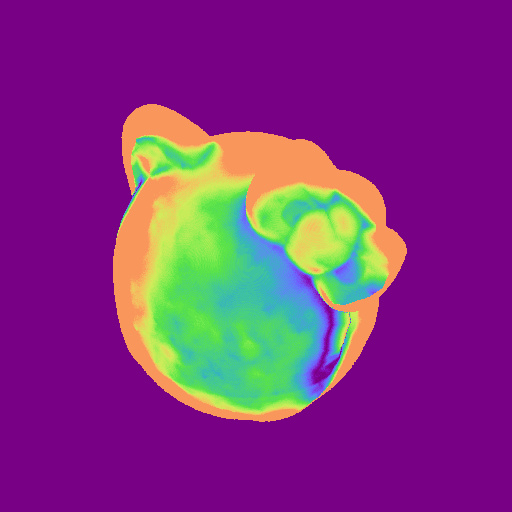}{80}{74}{80}{86}\\
   		\textit{kitty} & 0.0305 & 0.0192 & 0.0324 & \textbf{0.0010} \\[5pt]
   		\\
		\ablationTwo{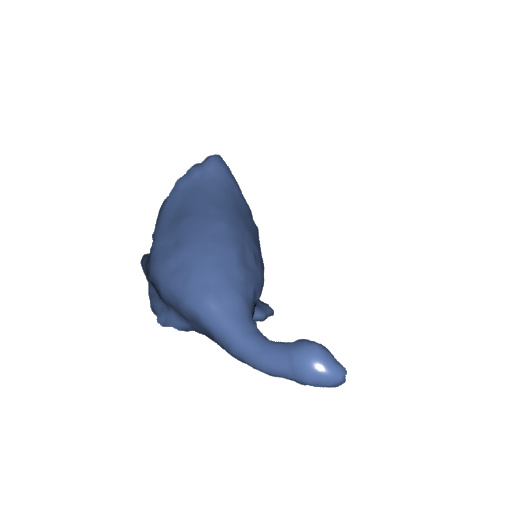}{115}{110}{115}{120}\\[5pt]
		\ablationTwoErr{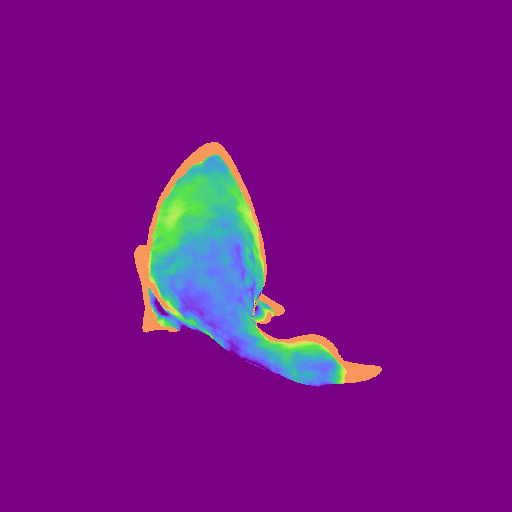}{115}{110}{115}{120}\\
		\textit{duck} & 0.0135 & 0.0079 & 0.0353 & \textbf{0.0011} \\[10pt]
   		\ablationTwo{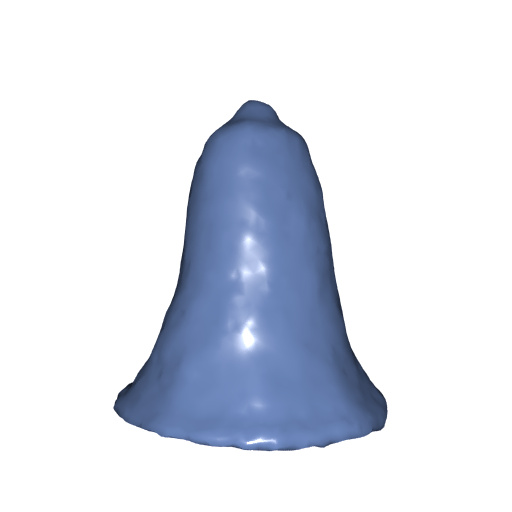}{57}{42}{57}{72}\\[5pt]
		\ablationTwoErr{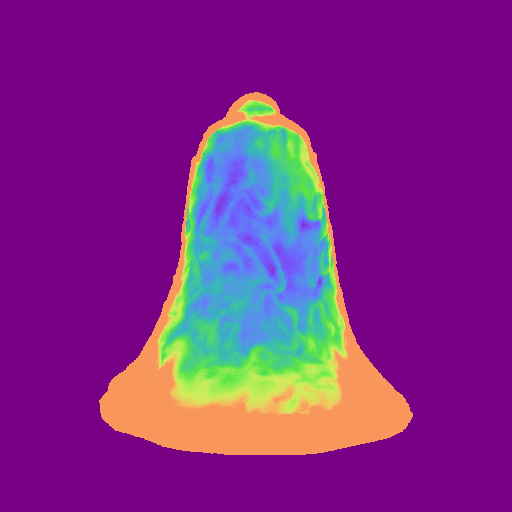}{57}{42}{57}{72}\\
		\textit{bell} & 0.0021 & 0.0197 & 0.0443 & \textbf{0.0005} \\[5pt]
  	\end{tabular}
  	\includegraphics[width=0.475\textwidth]{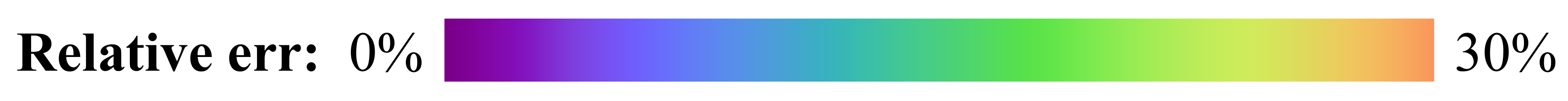}
    \caption{%
    	\small
    	\textbf{Comparison} with COLMAP~\cite{schoenberger2016mvs} and Kinect Fusion~\cite{newcombe2011kinectfusion} using synthetic inputs.
    	The \emph{COLMAP} results are generated using exact camera poses; the \emph{KF (High)} and \emph{KF (Low)} results are created using Kinect Fusion with high-resolution ground-truth depth and low-resolution noisy depth, respectively.
    	Our method, when using crude initializations given by \emph{KF (Low)} and RGB inputs, produces much more accurate geometries than \emph{COLMAP} and \emph{KF (High)}.
    	The number below each result indicates the average point-to-mesh distance.
    }
	\label{fig:ablation2_supp}
\end{figure*}

\paragraph*{Additional comparisons.}
We show in Figure~\ref{fig:ablation2_supp} additional comparisons between geometries reconstructed using COLMAP~\cite{schoenberger2016mvs}, Kinect Fusion~\cite{newcombe2011kinectfusion}, and our method.
Using coarse initializations provided with \emph{KF~(Low)}, our method consistently outperforms both \emph{COLMAP} and \emph{KF~(High)}.

\begin{figure*}[p]
	\centering
    \setlength{\resLen}{0.985in}
	\newcommand{\ablationOne}[5]{%
		\includegraphics[trim=#2 #3 #4 #5, clip, width=\resLen]{results/ablation1/#1/supp_init2.jpg} &
		\includegraphics[trim=#2 #3 #4 #5, clip, width=\resLen]{results/ablation1/#1/supp_softras2.jpg} &
		\includegraphics[trim=#2 #3 #4 #5, clip, width=\resLen]{results/ablation1/#1/supp_pytorch3d2.jpg} &
		\includegraphics[trim=#2 #3 #4 #5, clip, width=\resLen]{results/ablation1/#1/supp_mitsuba2.jpg} &
		\includegraphics[trim=#2 #3 #4 #5, clip, width=\resLen]{results/ablation1/#1/supp_nvdiffrast2.png} &
		\includegraphics[trim=#2 #3 #4 #5, clip, width=\resLen]{results/ablation1/#1/supp_ours2.jpg} &
		\includegraphics[trim=#2 #3 #4 #5, clip, width=\resLen]{results/ablation1/#1/supp_gt2.jpg}%
	}
	\newcommand{\ablationOneErr}[5]{%
		\includegraphics[trim=#2 #3 #4 #5, clip, width=\resLen]{results/ablation1/#1/supp_init_err.jpg} &
		\includegraphics[trim=#2 #3 #4 #5, clip, width=\resLen]{results/ablation1/#1/supp_softras_err.jpg} &
		\includegraphics[trim=#2 #3 #4 #5, clip, width=\resLen]{results/ablation1/#1/supp_pytorch3d_err.jpg} &
		\includegraphics[trim=#2 #3 #4 #5, clip, width=\resLen]{results/ablation1/#1/supp_mitsuba2_err.jpg} &
		\includegraphics[trim=#2 #3 #4 #5, clip, width=\resLen]{results/ablation1/#1/supp_nvdiffrast_err.png} &
		\includegraphics[trim=#2 #3 #4 #5, clip, width=\resLen]{results/ablation1/#1/supp_ours_err.jpg} &
	}
	\addtolength{\tabcolsep}{-5pt}
	\small
   	\begin{tabular}{ccccccc}
		\textsc{\scriptsize Init. mesh} &
		\textsc{\scriptsize SoftRas} &
		\textsc{\scriptsize Pytorch3D} &
		\textsc{\scriptsize Mitsuba 2} &
		\textsc{\scriptsize Nvdiffrast} &
		\textsc{\scriptsize Ours} &
		\textsc{\scriptsize Ground truth}\\
		\ablationOne{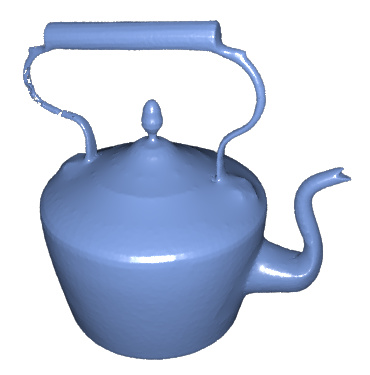}{75}{71}{75}{79}\\
		0.0411 & 0.0051 & 0.0072 & 0.0071 & 0.0013 & \textbf{0.0004} & \textit{kettle}\\
		\ablationOneErr{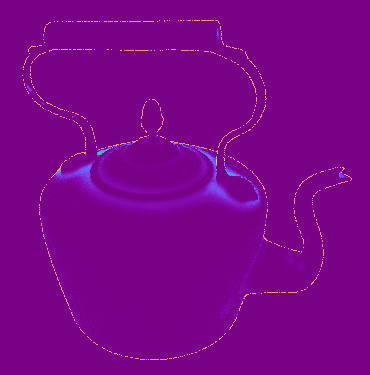}{75}{71}{75}{79}\\

		\ablationOne{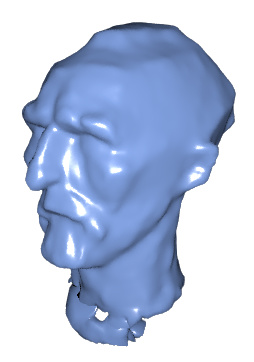}{75}{77}{75}{73}\\
		0.0115 & 0.0043 & 0.0091 & 0.0064 & 0.0022 & \textbf{0.0016} & \textit{head}\\
		\ablationOneErr{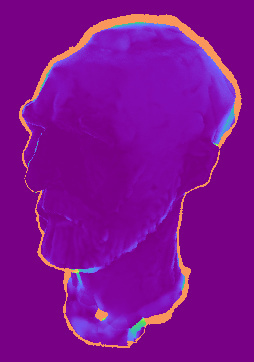}{75}{77}{75}{73}\\
		
		\ablationOne{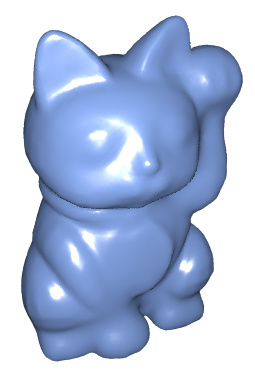}{75}{67}{75}{83}\\
		0.0878 & 0.0053 & 0.0066 & 0.0071 & 0.0023 & \textbf{0.0010} & \textit{maneki}\\
		\ablationOneErr{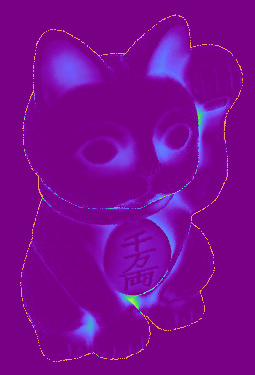}{75}{67}{75}{83}\\
		
	\end{tabular}
	\includegraphics[width=0.475\textwidth]{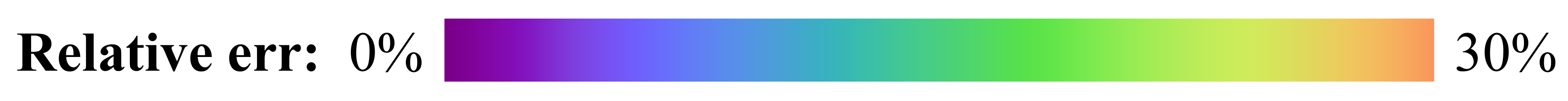}
    \caption{
    	\small
    	\textbf{Comparison} with SoftRas~\cite{liu2019soft}, PyTorch3D~\cite{ravi2020accelerating}, Mitsuba 2~\cite{nimier2019mitsuba} and Nvdiffrast~\cite{Laine2020diffrast}.
    	We render all reconstructed geometries using Phong shading and visualize depth errors (wrt. the ground-truth geometry).
    	Initialized with the same mesh (shown in the left column), optimizations using gradients obtained with SoftRas and PyTorch3D tend to converge to low-quality results due to gradient inaccuracies caused by soft rasterization.
    	Mitsuba~2, a ray-tracing-based system, also produces visible artifacts due to biased gradients resulting from an approximated reparameterization~\cite{loubet2019reparameterizing}. Nvdiffrast is using multisample analytic antialiasing method to provide reliable visibility gradients, which yields better optimization result overall. 
    	When using gradients generated with our differentiable renderer, optimizations under identical configurations produce results closely resembling the targets.
    	The number below each result indicates the average point-to-mesh distance.
    }
    \label{fig:ablation1}
\end{figure*}

We further demonstrate the importance of having accurate geometric gradients in Figure~\ref{fig:ablation1}.
When using identical initialization and optimization configurations, gradients generated by our differentiable renderer (depicted in \S 3 of the paper) can yield high-quality optimization results.
Using biased gradient estimates given by existing methods like SoftRas~\cite{liu2019soft}, PyTorch3D~\cite{ravi2020accelerating}, Mitsuba~2~\cite{nimier2019mitsuba} and Nvdiffrast~\cite{Laine2020diffrast}, on the other hand, produces reconstructions with consistently worse qualities.

\section{Additional results}
\label{sec:recon}
\begin{figure*}[p]
 	\centering
   	\setlength{\resLen}{1.32in}
   	\newcommand{\resultsSyn}[1]{%
   		\begin{overpic}[width=\resLen]{results/synthetic/#1/KF2.jpg}
   			\put(2, 88) {\small\textcolor{black}{\bfseries #1}}
   		\end{overpic}
   		&
		\includegraphics[width=\resLen]{results/synthetic/#1/novel_tar2.jpg} &
		\includegraphics[width=\resLen]{results/synthetic/#1/novel_fit2.jpg} &
		\includegraphics[width=\resLen]{results/synthetic/#1/normal2.jpg}%
	}
   	\addtolength{\tabcolsep}{-5pt}
	\begin{tabular}{cccc}
		\textsc{\scriptsize Init. mesh} &
		\textsc{\scriptsize GT (point-light)} &
		\textsc{\scriptsize Ours (point-light)} &
		\textsc{\scriptsize Ours (normal)}\\
		\resultsSyn{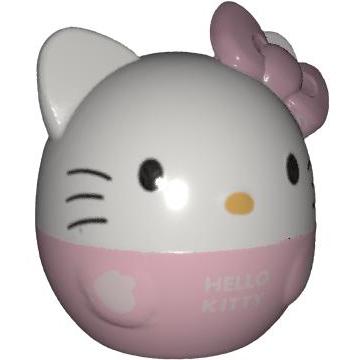}\\[-2pt]
		\resultsSyn{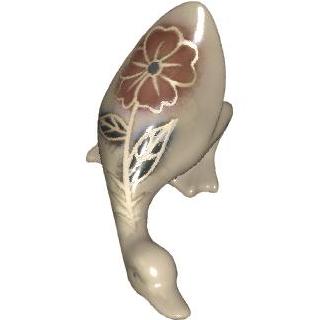}\\[-2pt]
		\resultsSyn{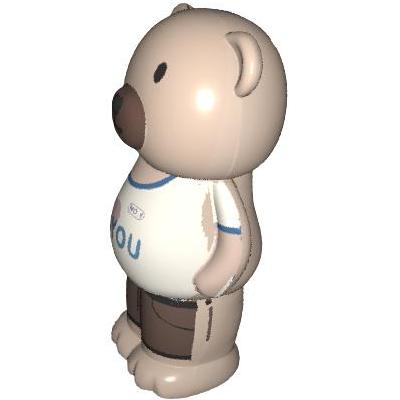}\\[-2pt]
		\resultsSyn{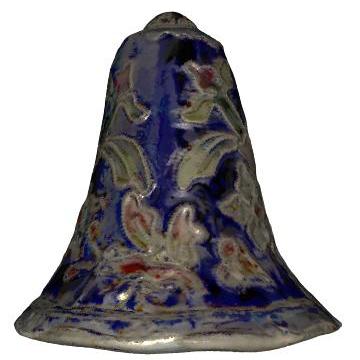}\\[-2pt]
		\resultsSyn{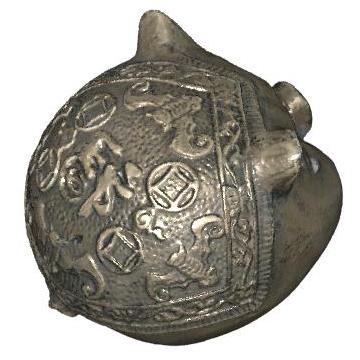}
	\end{tabular}
    \caption{%
    	\small
    	\textbf{Reconstruction results} using synthetic inputs:
    	We obtain the initial meshes (in the left column) using Kinect Fusion~\cite{newcombe2011kinectfusion} with low-resolution ($48 \times 48$) and noisy depths.
    	The detailed normal variations in our models emerge entirely from the reconstructed geometries (and no displacement or normal mapping is used).
    }
	\label{fig:main_syn1}
\end{figure*}

\begin{figure*}[p]
 	\centering
	\setlength{\resLen}{1.32in}
    	\newcommand{\resultsReal}[3]{%
		\begin{overpic}[width=\resLen]{results/real/#1/supp_img#2.jpg}
			\put(2, 4) {\small\textcolor{white}{\bfseries #1}}
		\end{overpic}
		&
		\includegraphics[width=\resLen]{results/real/#1/supp_our_#3.jpg} &
		\includegraphics[width=\resLen]{results/real/#1/supp_normal_#3.jpg} &
		\includegraphics[width=\resLen]{results/real/#1/supp_novel_1.jpg} &
		\includegraphics[width=\resLen]{results/real/#1/supp_novel_2.jpg}%
	}
   	\addtolength{\tabcolsep}{-5pt}
	\begin{tabular}{ccccc}
		\textsc{\scriptsize Photo. (novel)} &
		\textsc{\scriptsize Ours} &
		\textsc{\scriptsize Normal} &
		\textsc{\scriptsize Ours (env. map 1)} &
		\textsc{\scriptsize Ours (env. map 2)} 
		\\
		\resultsReal{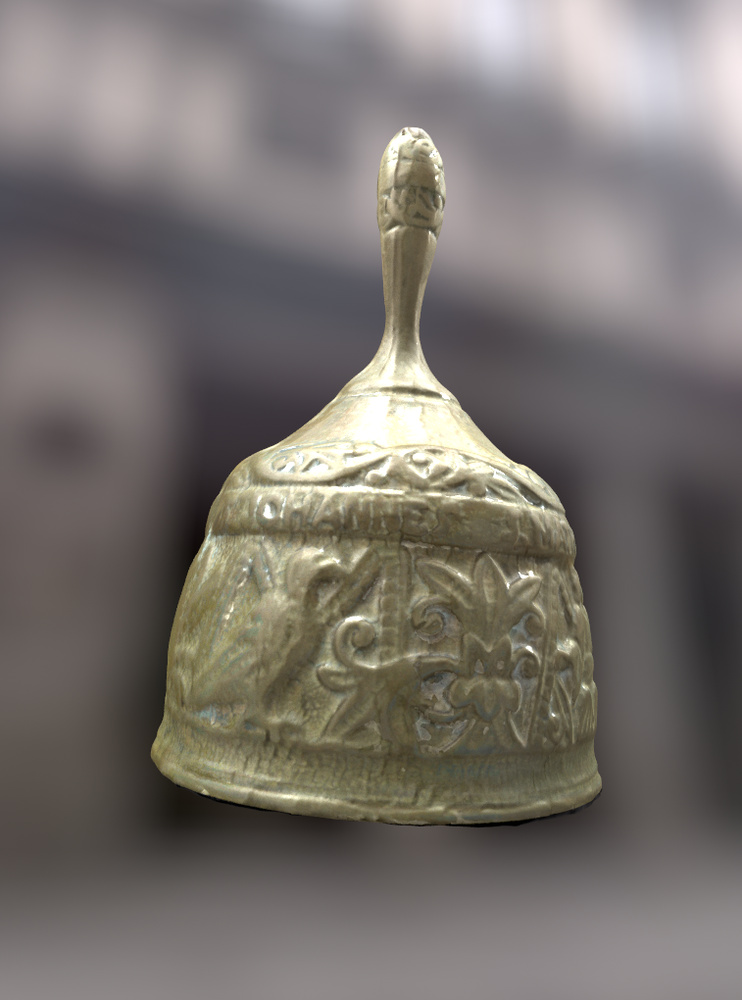}{0229}{229}\\[-2pt]
		\resultsReal{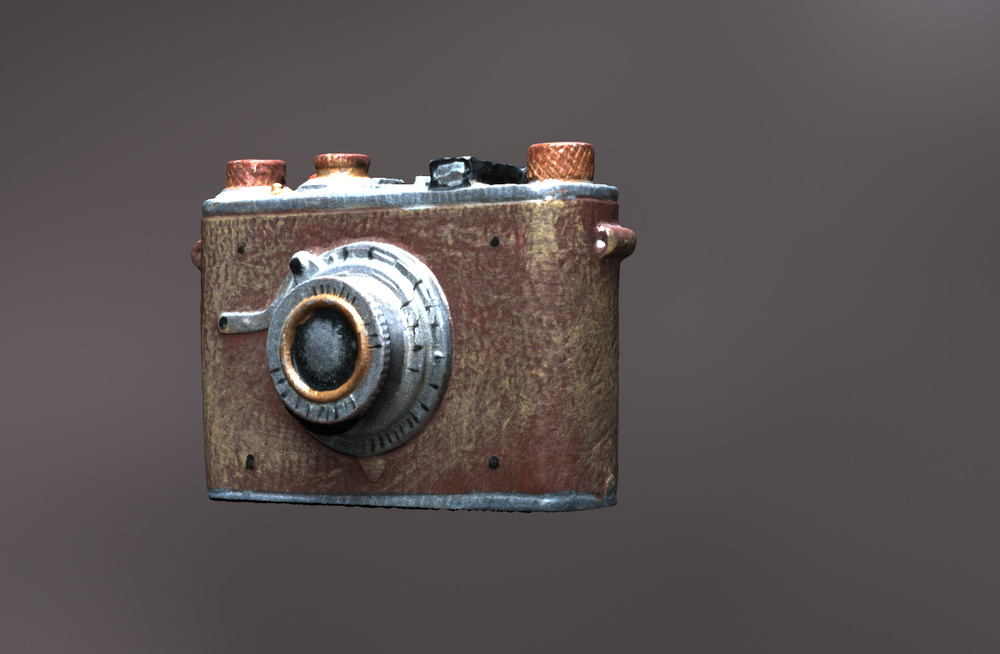}{0063}{63}\\[-2pt]
		\resultsReal{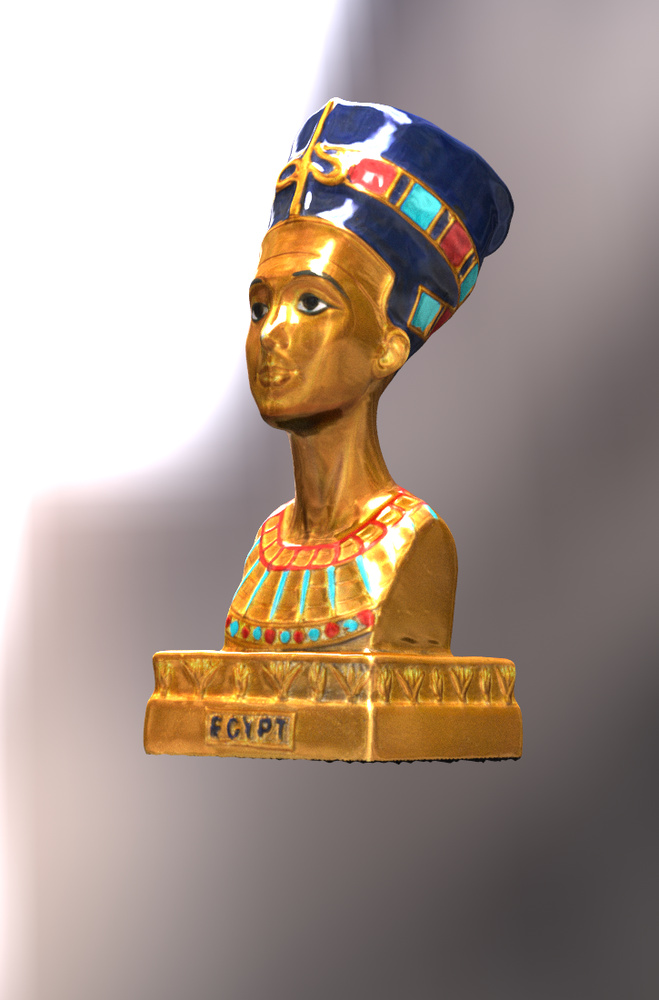}{0001}{1}\\[-2pt]
		\resultsReal{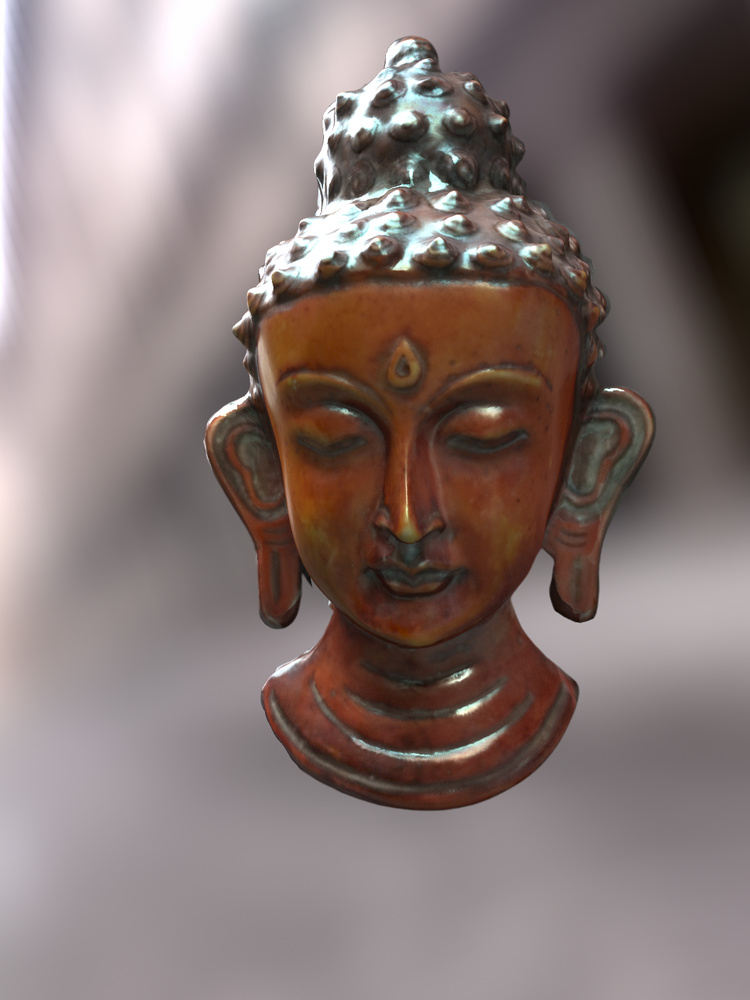}{0072}{72}\\[-2pt]
		\resultsReal{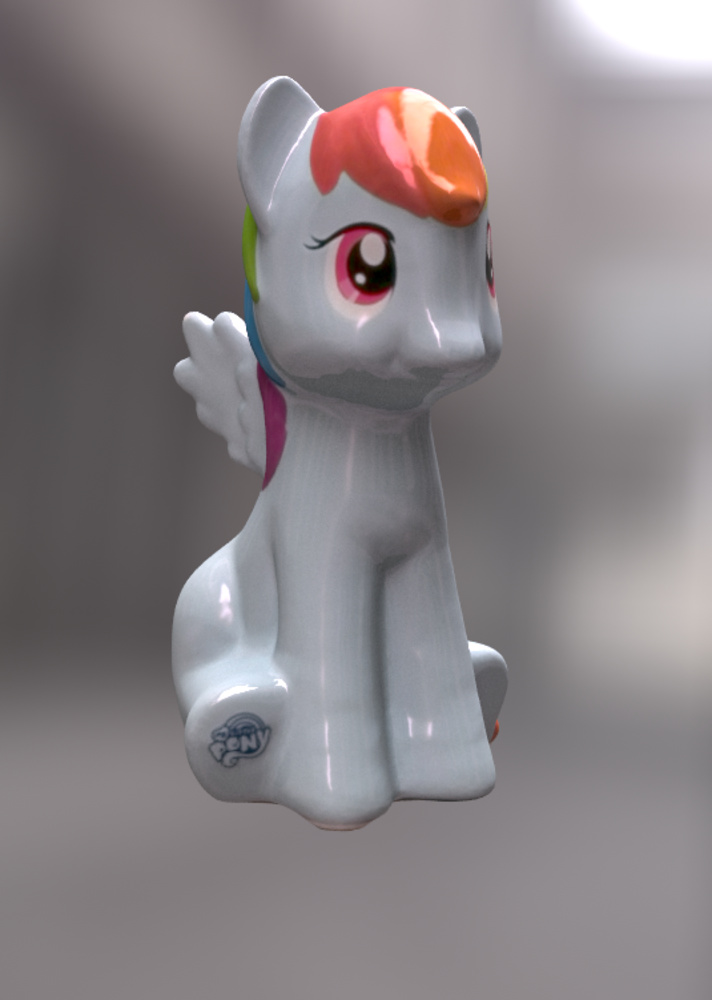}{0165}{165}
	\end{tabular}
    \caption{%
    	\small
    	\textbf{Reconstruction results} of real-world objects: Similar to the synthetic examples in Figure~\protect\ref{fig:main_syn1}, our technique recovers detailed object geometries and reflectance details, producing high-quality results under novel viewing and illumination conditions.
    }
	\label{fig:main2_supp}
\end{figure*}

\paragraph*{Reconstruction results.}
We show more reconstruction results of synthetic objects in Figure~\ref{fig:main_syn1} and real ones in Figure~\ref{fig:main2_supp}.

\begin{figure*}[p]
	\centering
	\includegraphics[width=1.8\columnwidth]{results/3d_modeling.jpg}\\[1pt]
	\includegraphics[height=0.7148\columnwidth]{results/teaser.jpg}
	\includegraphics[height=0.7148\columnwidth]{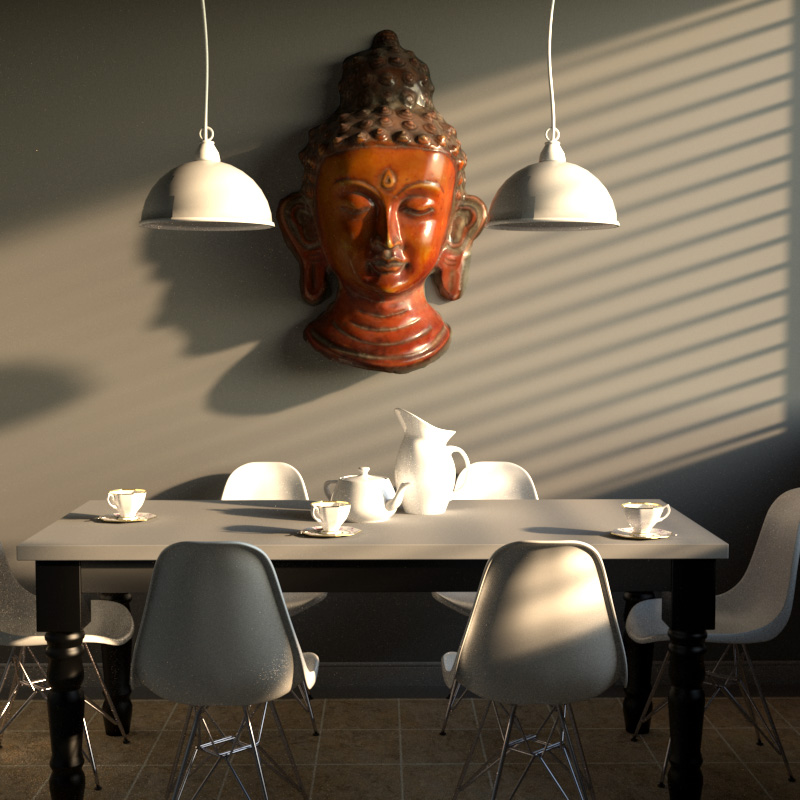}
	\caption{%
		\small
		\textbf{Physics-based renderings} of our models (i.e., everything on the tables in the top and bottom-left images; the buddha model on the wall in the bottom-right image) within complex virtual scenes.
	}
	\label{fig:app_cg}
\end{figure*}

\begin{figure*}[p]
	\centering
	\includegraphics[width=1.8\columnwidth]{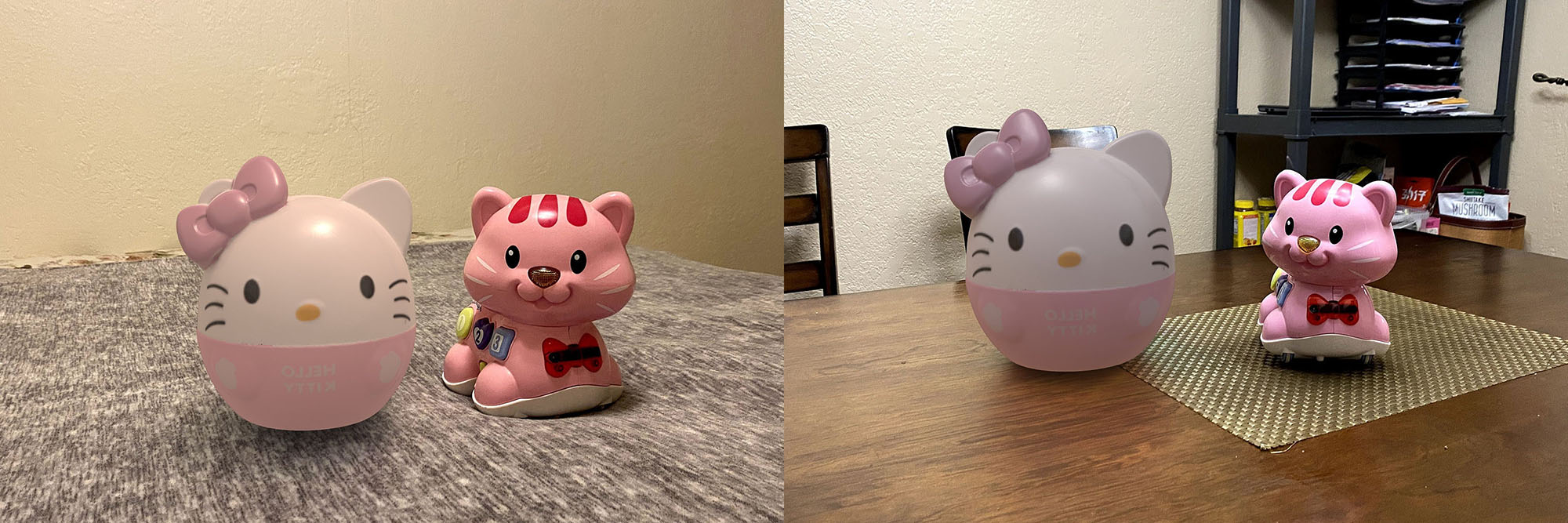}
	\caption{%
		\small
		\textbf{AR object insertion:}
		Our reconstructed \emph{kitty} model (left) inserted into a real scene; the object on the right is real.
	}
	\label{fig:app_ar}
\end{figure*}

\paragraph*{Applications.}
Since our reconstructions use standard mesh-based representations, they can be easily used in a wide range of applications.
In Figure~\ref{fig:app_cg}, we show physics-based renderings of our reconstructed models in complex virtual scenes.
Figure~\ref{fig:app_ar} shows augmented reality (AR) examples where our models are inserted into two real scenes.

\begin{figure*}[p]
 	\centering
   	\setlength{\resLen}{1.32in}
   	\newcommand{\resultsSyn}[1]{%
		\includegraphics[width=\resLen]{results/synthetic/#1/envmap1_tar2.jpg} &
		\includegraphics[width=\resLen]{results/synthetic/#1/envmap1_fit2.jpg} &
		\includegraphics[width=\resLen]{results/synthetic/#1/envmap2_tar2.jpg} &
		\includegraphics[width=\resLen]{results/synthetic/#1/envmap2_fit2.jpg}%
	}
   	\addtolength{\tabcolsep}{-5pt}
	\begin{tabular}{cccc}
		\textsc{\scriptsize GT (env. map 1)} &
		\textsc{\scriptsize Ours (env. map 1)} &
		\textsc{\scriptsize GT (env. map 2)} &
		\textsc{\scriptsize Ours (env. map 2)}\\
		\resultsSyn{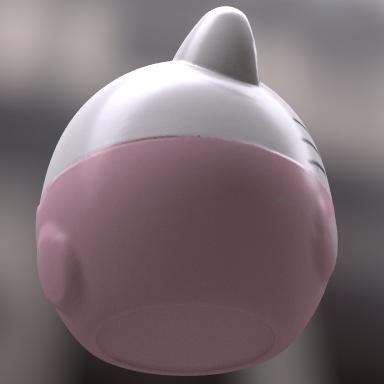}\\[-2pt]
		\resultsSyn{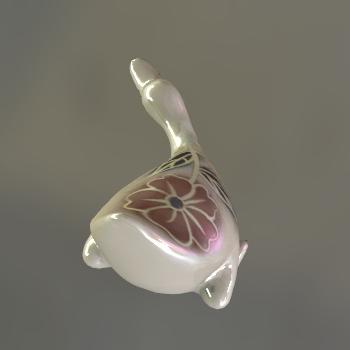}\\[-2pt]
		\resultsSyn{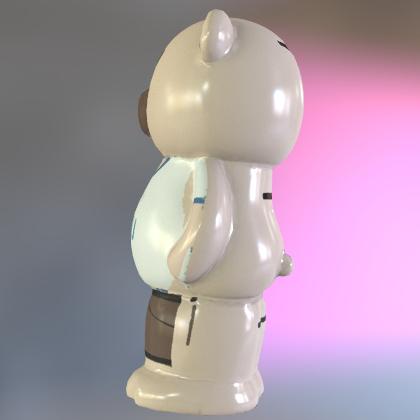}\\[-2pt]
		\resultsSyn{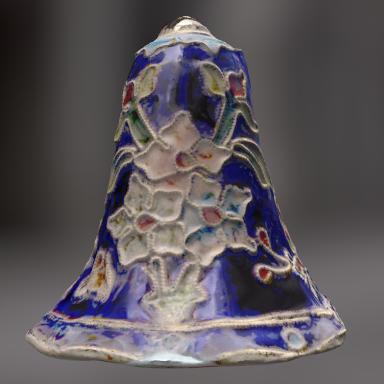}\\[-2pt]
		\resultsSyn{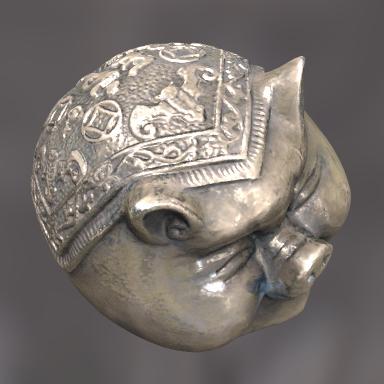}
	\end{tabular}
    \caption[]{%
    	\small
    	\textbf{Reconstruction results} using synthetic inputs: Re-renderings of the reconstruction results from Figure~\protect\ref{fig:main_syn1} under environmental illuminations.
    }
	\label{fig:main_syn1_env}
\end{figure*}

\clearpage
\begingroup
	\small
	\bibliographystyle{eg-alpha-doi}
	\bibliography{egbib}
\endgroup